%% file: main.tex
\documentclass[acmsmall]{acmart}

\renewcommand\footnotetextcopyrightpermission[1]{} % removes footnote with conference information in first column
\setcopyright{acmcopyright}
\copyrightyear{2026}
\acmYear{2026}
\acmDOI{}

\usepackage{subfig}
\usepackage{colortbl}
\definecolor{inteins}{RGB}{128,179,255}
\definecolor{intzwei}{RGB}{42,127,255}
\definecolor{intdrei}{RGB}{0,85,212}
\definecolor{intvier}{RGB}{255,143,143}

\usepackage{multirow}
\usepackage{bm}
\usepackage{xcolor}
\newcommand{\minorrevise}[1]{#1}
\newcommand{\revise}[1]{#1}

\begin{document}

\title[From Data Alignment to Parameter Generation: Few-Shot Cross-City Traffic Prediction]{From Data Alignment to Parameter Generation: Few-Shot Cross-City Traffic Prediction with Conditional Diffusion}

\author{Shengchun Wang}
\affiliation{%
  \institution{Hunan Normal University}
  \city{Changsha}
  \country{China}}
\email{scwang@hunnu.edu.cn}

\author{Yuhang Lin}
\affiliation{%
  \institution{Hunan Normal University}
  \city{Changsha}
  \country{China}}
\email{yuhang_lin@hunnu.edu.cn}

\author{Tong Li}
\affiliation{%
  \institution{Hunan University}
  \city{Changsha}
  \country{China}}
\email{tliay@hnu.edu.cn}

\author{Pengyuan Zhou}
\affiliation{%
  \institution{Aarhus University}
  \city{Aarhus}
  \country{Denmark}}
\email{pengyuan.zhou@ece.au.dk}

\author{Andy Chi Lok TAI}
\affiliation{%
  \institution{The Hong Kong Polytechnic University}
  \city{Hong Kong}
  \country{China}}
\email{andy.tai@cpce-polyu.edu.hk}

\author{Wenjie Chen}
\affiliation{%
  \institution{Central South University of Forestry and Technology}
  \city{Changsha}
  \country{China}}
\email{t20142193@csuft.edu.cn}

\author{Zhu Xiao}
\affiliation{%
  \institution{Hunan University}
  \city{Changsha}
  \country{China}}
\email{zhxiao@hnu.edu.cn}

\author{Zhu Han}
\affiliation{%
  \institution{University of Houston}
  \city{Houston}
  \country{USA}}
\affiliation{%
  \institution{Kyung Hee University}
  \city{Seoul}
  \country{South Korea}}
\email{hanzhu22@gmail.com}

\renewcommand{\shortauthors}{S. Wang et al.}

\begin{abstract}
Accurate mobile traffic forecasting is crucial for network service management, supporting proactive capacity provisioning, congestion mitigation, and energy-efficient operation. However, many cities and regions are data-scarce (e.g., newly deployed or poorly instrumented areas), thus making traffic prediction a few-shot learning problem and limiting the practicality of standard data-driven models. Existing cross-city solutions primarily rely on data-space alignment to match traffic distributions between data-rich and data-scarce cities, but such alignment is often brittle under real-world distribution shifts, measurement noise, and city-specific bias. To address these limitations, we propose a Conditional Diffusion-based Parameter Generation (CDPG) framework. Instead of aligning raw data, CDPG learns the conditional distribution of high-performing predictor parameters for source cities and directly generates node-specific model parameters for the target city. To support reliable adaptation for network traffic prediction, CDPG conditions parameter generation on multi-modal spatio-temporal context. Specifically, CDPG captures spatial environment cues by combining satellite imagery with large language model (LLM)-derived semantic descriptions, and summarizes complex temporal behaviors via unsupervised pattern clustering. These contextual components guide CDPG in producing parameters tailored to each base station's operational context, thereby improving robustness under cross-city and cross-region shifts. Extensive experiments on three large-scale datasets show that CDPG consistently outperforms state-of-the-art cross-city few-shot baselines by up to 9.72\% on standard forecasting metrics and remains robust for few-shot multi-step prediction. The code is available at \href{https://github.com/LingoodBoy/CDPG}{https://github.com/LingoodBoy/CDPG}.
\end{abstract}



\maketitle

\input{Introduction}  %  1.5 page	
\input{Preliminary} % 0.75 page
\input{Methods} % 1 page
\input{Experiments} % 2 pages
\input{Related} %  1.5 page

\section{Conclusion}
\label{sec:conclusion} 

In this paper, we proposed CDPG, a parameter-space generative framework designed to tackle the problem of few-shot cross-city network traffic prediction. Unlike traditional methods that align divergent data distributions, CDPG introduces a paradigm shift by learning the conditional distribution of optimal neural network parameters via a diffusion model. CDPG enables the robust initialization of specialized, lightweight predictors tailored to the unique spatio-temporal context of each base station. To support this generation, we designed a multi-modal conditioning mechanism that integrates LLM-enhanced spatial semantics and clustering-based temporal patterns. 
Extensive experiments on three real-world datasets demonstrate that CDPG significantly outperforms state-of-the-art baselines for few-shot cross-city prediction, achieving improvements of up to 9.72\% on MAE, RMSE, and $R^2$. Moreover, CDPG instantiates a lightweight predictor for each base station with only 6K parameters, confirming its effectiveness and robustness for network traffic prediction.

% \section*{Acknowledgments}
% This research has been supported in part by the National Key Research and Development Program of China under Grant 2020YFA0711403; in part by the National Natural Science Foundation of China under Grant 61972223, Grant U1936217, and Grant 61971267; in part by the Guoqiang Institute under Grant 2021GQG1005; in part by the International Postdoctoral Exchange Fellowship Program (Talent-Introduction Program) under Project YJ20210274; in part by the China Postdoctoral Science Foundation under Project 2022M721891; and in part by the Academy of Finland under Project 319669, Project 319670, Project 325570, Project 326305, Project 325774, and Project 335934.

\bibliographystyle{unsrt}
\bibliography{Ref}
\newpage

\end{document}

%% file: Introduction.tex
\section{Introduction}
\label{sec:Introduction}
Mobile networks have evolved into programmable service infrastructures, where operators must continuously meet service-level objectives such as throughput, latency, availability, and energy efficiency. Within this service management loop, network traffic prediction is a fundamental capability~\cite{xie2024m, gong2023scalable, zhang2026lsdm, xiao2025llm}. Accurate forecasts enable proactive resource orchestration (e.g., bandwidth provisioning and spectrum scheduling), service assurance (e.g., congestion prevention and load balancing), and energy-aware operation (e.g., base-station sleep control), thereby improving user experience while reducing operational cost and carbon footprint~\cite{su2025gennet, li2023carbon, huang2023safe, sun2024dynamic, bega2020deepcog}.
\revise{ %R3-C4
These management actions are increasingly automated and time-sensitive, making prediction robustness as important as pointwise accuracy.
}
Driven by the availability of large-scale network telemetry, data-driven forecasting has become the mainstream paradigm~\cite{gong2025sttf, zhang2021graph, yang2023transforms, li2024mobile}. Prior work ranges from classical statistical models~\cite{hong2012application, zhang2022cityneuro} to deep spatio-temporal architectures~\cite{li2019privacy, chai2024knowledge, liu2026multi}, including graph neural networks (GNNs)~\cite{li2022dmgan, gu2023spatial, gong2023empowering}, which jointly learn the spatial dependence among network elements and the temporal dynamics of traffic series. Despite their success, these methods typically assume that a target network has sufficient and representative historical data for training or fine-tuning. In practical service deployments, however, this assumption is often violated: new rollouts, newly upgraded sites, incomplete instrumentation, privacy and governance constraints, and measurement failures can all yield data-scarce cities or regions with skewed or noisy observations~\cite{liu2024deep}. In such cases, directly training high-capacity predictors is prone to overfitting and poor generalization, undermining their reliability for service management decisions.

A natural direction is cross-city transfer: leverage data-rich source cities to improve prediction in a data-scarce target city under a few-shot setting~\cite{aouedi2025deep}. Existing cross-city methods largely follow a data alignment or model adaptation paradigm. Early approaches transfer knowledge by matching regions with similar patterns using clustering and heuristic similarity measures~\cite{zhang2019deep, dridi2021transfer}. Later work performs more explicit distribution alignment (e.g., adversarial learning) to reduce domain shift~\cite{wu2021deep, zhang2023deep}. \revise{ % R1-C2、R3-C5
Meta-learning methods learn transferable initializations from multiple cities for fast target-city adaptation~\cite{yao2019learning}. Foundation-model-based solutions improve generalization through large-scale pre-training and lightweight adaptation, such as prompt tuning~\cite{hu2024prompt}. Continual and cross-domain spatio-temporal learning methods exploit shared knowledge across evolving tasks or domains~\cite{yi2024get, liu2025synevo}.
} 
While these approaches can be effective, they share a common limitation in network service management. \revise{ % R1-C2
These lines of work mainly rely on adapting data representations, shared model states, or domain-level knowledge from observed spatio-temporal samples.
} 
However, raw traffic observations are strongly affected by cross-city differences in infrastructure, population mobility, sensing quality, and operational policies, which induce substantial distribution shifts, noise, and bias. When the raw distributions are complex to align (or alignment is imperfect), transfer may become brittle or even negative~\cite{wu2021deep, chai2025uomo}, a problematic property for service assurance workflows that require stable performance under heterogeneity.

Motivated by this limitation, we advocate a paradigm shift from data alignment to parameter generation. Instead of forcing data from different cities to match, we aim to directly generate a well-performing predictor for the target city by modeling the relationship between a network element's context and the predictor parameters it should use.
\revise{ % R1-C2
Prior parameter-generation methods, such as hypernetwork-style generators and dynamic parameter learners, have shown that model parameters can be conditioned on task or location context~\cite{ha2016hypernetworks, pan2019urban, cirstea2022towards}. More recently, diffusion-based neural parameter generation has been explored for general spatio-temporal few-shot learning~\cite{yuan2024spatio}. However, these methods do not fully address few-shot cross-city mobile traffic prediction, where each base station may exhibit distinct physical surroundings, functional semantics, and temporal traffic patterns.
} 
\revise{ % r3-c5
This parameter-space view is useful because model parameters compactly represent forecasting behavior. Generating parameters conditioned on context can reduce reliance on noisy target observations and support rapid adaptation from a few target samples.
}
Achieving few-shot cross-city parameter generation for traffic prediction introduces three challenges.
\begin{itemize}
\item \textbf{Cross-city and Cross-region Distribution Shifts.} 
Real network telemetry is affected by heterogeneous deployment conditions and measurement imperfections, leading to shifts not only across cities but also within a city. A practical solution must remain robust under such multi-level shifts.
\item \textbf{Implicit Spatial Dependencies.} 
Regions that are geographically separated may share similar urban functions or physical layouts (e.g., transportation hubs, residential blocks, business districts), and therefore exhibit similar traffic behaviors. These semantics are often absent from raw traffic records and difficult to exploit with purely numeric time-series transfer.
\item \textbf{Complex Temporal Dynamics.}  Mobile traffic combines periodic patterns (e.g., diurnal/weekly cycles) with irregular bursts (events, anomalies, stochastic user behavior). Distilling transferable temporal representations from a few target samples is non-trivial.
\end{itemize}

To address the above challenges, we propose a \textbf{C}onditional \textbf{D}iffusion-based \textbf{P}arameter \textbf{G}eneration (CDPG) framework for few-shot cross-city mobile traffic prediction. CDPG learns, from data-rich source cities, the conditional distribution of high-performing neural parameters and then generates base-station-specific predictor parameters for a data-scarce target city. A diffusion model iteratively denoises random noise into functional parameters, leveraging its refinement capability to mitigate sensitivity to noisy observations and distribution shifts, thereby addressing the cross-city and cross-region shift challenge. To improve conditioning and capture latent similarities that matter for service management, CDPG incorporates multi-modal spatio-temporal context. For the spatial context, CDPG combines satellite imagery with large language model (LLM)-derived semantic descriptions to encode both physical layout cues and higher-level functional semantics, and aligns them via a bidirectional contrastive learning scheme. For the temporal context, CDPG performs unsupervised clustering to construct a compact taxonomy of representative traffic patterns, which guides parameter generation to handle heterogeneous dynamics under few-shot supervision. Together, these designs generate parameters tailored to each network element's spatio-temporal environment, enabling reliable predictions for downstream orchestration and assurance tasks.

Our contributions are summarized as follows:
\begin{itemize}

\item 
\revise{ % R3-C4
We investigate few-shot cross-city mobile traffic prediction and propose a parameter-space generative paradigm that shifts from data-space alignment to parameter generation.
}
Our paradigm bypasses substantial shifts in the distribution of mobile traffic data by directly learning the conditional distribution of optimal neural parameters, enabling robust generation of specialized neural parameters for predictors in data-scarce cities.

\item We propose the CDPG framework, which employs a Diffusion-based neural parameter generation paradigm to generate specialized parameters, effectively bypassing substantial data distribution shifts by modeling the intrinsic environment-parameter mapping. To support this, CDPG uses LLM-enhanced spatial context extraction to capture latent functional dependencies that are invisible in raw mobile traffic data by aligning satellite imagery with semantic descriptions. Additionally, CDPG leverages pattern-guided temporal context extraction to disentangle complex dynamics into discrete 
patterns, ensuring the generated predictors are accurately tailored to heterogeneous temporal trends.

\item We conduct extensive experiments on few-shot cross-city prediction tasks across three large-scale mobile traffic datasets, demonstrating that CDPG significantly outperforms state-of-the-art cross-city few-shot prediction baselines by up to 9.72\% across mean absolute error (MAE), root mean squared error (RMSE), and $R^2$ metrics and exhibits remarkable robustness in few-shot multi-step mobile traffic prediction tasks. Our results validate the effectiveness of the parameter-space paradigm and the proposed multi-modal conditioning for rapid adaptation in the data-scarce city.
\end{itemize}

The remainder of this paper is organized as follows. 
We begin by presenting the preliminaries in Section~\ref{sec:Preliminary}. Next, we elaborate on the proposed CDPG framework, detailing the parameter-space generative mechanism and the spatio-temporal condition designs in Section~\ref{sec:Method}. Following the methodology, we present the experimental settings and extensive evaluation results in Section~\ref{sec:Experiment}. Finally, after discussing related work in Section~\ref{sec:related_work}, we conclude the paper in Section~\ref{sec:conclusion}.

%% file: Preliminary.tex
\section{Preliminary}
\label{sec:Preliminary}

\subsection{Neural Parameter Generation}
\label{sec:Preliminary_parameter_generation}
\begin{figure}[tb]
\centering
\includegraphics[width=0.8\linewidth]{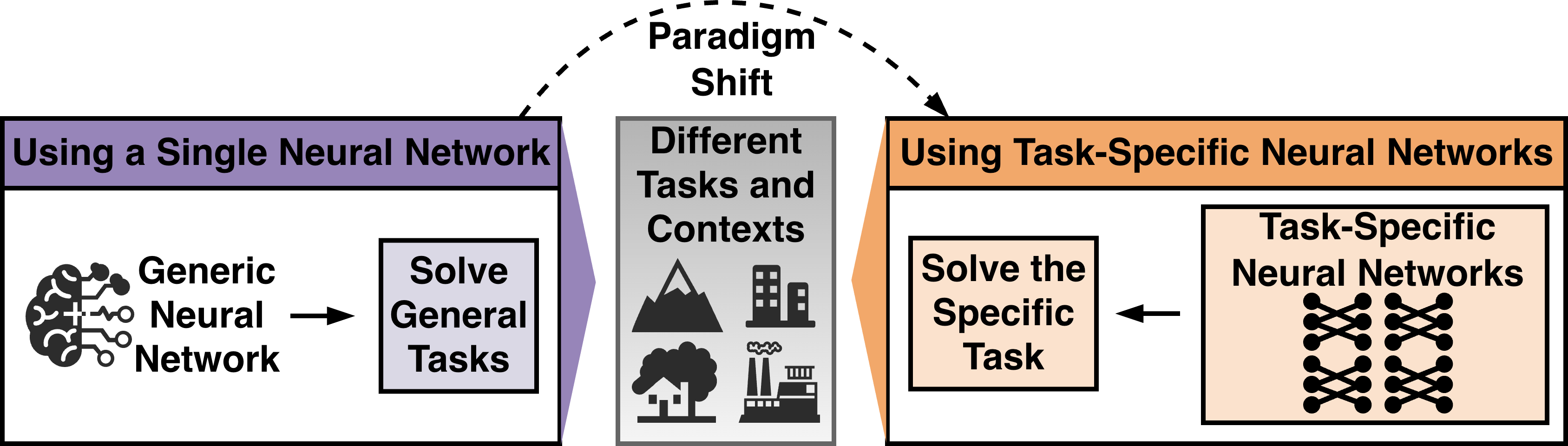} 
\Description{A comparison diagram showing the paradigm shift.}
\vspace{-2mm}
\caption{Paradigm shift from a generic neural network to task-specific neural networks for different tasks.}
\label{fig:paradign_shift}
\end{figure}

Neural Parameter Generation (NPG) represents a fundamental paradigm shift from traditional optimization-based learning to model parameter generation~\cite{ha2016hypernetworks, alaluf2022hyperstyle, wang2024neural}, as illustrated in \figurename~\ref{fig:paradign_shift}.
Conventionally, deep learning methods learn a fixed set of parameters $\theta$ via gradient descent to minimize a loss function over a global dataset.
However, data distributions and task-specific contexts often vary significantly across different tasks. A shared $\theta$ typically fails to capture such localized patterns and suffers from poor generalization, especially in data-scarce scenarios. 

To address these challenges, NPG employs a hypernetwork parameterized by $\phi$ to generate the target task network's parameters $\theta$. Given a condition $c$ that encodes the task-specific context (e.g., regional attributes), the parameter generator $\mathcal{G}_\phi$ produces $\theta = \mathcal{G}_\phi(c)$. The generated parameters constitute a task-specific parameter set for the target tasks, thereby better capturing task-specific context~\cite{li2023gpt, pan2020spatio}.

\subsection{Conditional Diffusion Models}
\label{sec:Preliminary_diffusion}
Conditional diffusion models are based on probability theory and are effective at modeling the conditional distribution of complex data given auxiliary information~\cite{sohl2015deep, li2025generative, croitoru2023diffusion}. We employ conditional diffusion models to generate neural parameters tailored to a given spatio-temporal condition $c$. In our setting, the data are neural parameter vectors $\theta_0 \in \mathbb{R}^D$, and we aim to model their conditional distribution $p_\gamma(\theta_0 \mid c)$.
Formally, this process consists of a fixed forward process and a learnable reverse process. The forward process is a Markov chain that gradually perturbs $\theta_0$ into Gaussian noise:
\begin{equation}
    \theta_k = \sqrt{\bar{\alpha}_k}\,\theta_0 + \sqrt{1-\bar{\alpha}_k}\,\epsilon, \quad
    \epsilon \sim \mathcal{N}(0, I),
\end{equation}
where $\{\alpha_k\}_{k=1}^T$ is a predefined noise schedule and $\bar{\alpha}_k = \prod_{i=1}^k \alpha_i$. The reverse process denoises from $\theta_T$ back to $\theta_0$ using a noise-prediction network $\epsilon_\gamma(\theta_k, c, k)$ that is conditioned on $c$. The training objective follows:
\begin{equation}
    \min_{\gamma} \mathcal{L}(\gamma) = \min_{\gamma} \mathbb{E}_{\theta_0,\,\epsilon,\,k,\,c}
    \big[ \big\| \epsilon - \epsilon_\gamma(\theta_k, c, k) \big\|_2^2 \big],
    \label{eq:prelim_loss}
\end{equation}
where $\theta_k$ is obtained from $\theta_0$ and $\epsilon$ via the forward process. At inference time, we first sample $\theta_T \sim \mathcal{N}(0, I)$ and iteratively apply the conditional denoising update:
\begin{equation}
    \theta_{k-1} = \frac{1}{\sqrt{\alpha_k}}
    \left( \theta_k - \frac{1 - \alpha_k}{\sqrt{1 - \bar{\alpha}_k}}\, \epsilon_\gamma(\theta_k, c, k) \right)
    + \sigma_k \mathbf{z},
    \label{eq:prelim_sampling}
\end{equation}
where $\mathbf{z} \sim \mathcal{N}(0, I)$ and $\sigma_k$ is determined by the variance schedule. The final sample $\theta_0$ serves as a generated neural parameter set adapted to the condition $c$, which is then used to initialize the specified predictor.

\subsection{Lightweight Graph Transformer Predictor}
\label{sec:Preliminary_predictor}
Inspired by the ability of the Transformer architecture to capture long-range spatio-temporal dependencies~\cite{hu2022citywide, gu2023spatial, kougioumtzidis2025mobile}, we leverage a Lightweight Graph Transformer as the base predictor for mobile traffic. The model represents network elements as nodes in a graph $G=(V, E)$, processing the historical traffic sequence of the target node's top-$K$ neighbors, denoted as $X \in \mathbb{R}^{T \times K \times 1}$. It outputs the prediction value $\hat{Y}$ for the target network element. Neighboring network elements often exhibit correlated traffic patterns due to shared urban contexts, and so we use a spatial attention mechanism ($\operatorname{Attn}_s$) to capture these correlations, while using a temporal attention mechanism ($\operatorname{Attn}_t$) for long-range dependencies. To balance modeling power and parameter efficiency, we adopt a single-head self-attention architecture. The hidden representation $H$ is computed as
\begin{equation}
H = \operatorname{Attn}_t\big( \operatorname{Attn}_s(X W_s) W_t \big),
\label{eq:prelim_graphtransformer}
\end{equation}
where $\operatorname{Attn}_s$ and $\operatorname{Attn}_t$ aggregate spatial and temporal features, and $W_s, W_t$ are learnable matrices. The traffic prediction is
\begin{equation}
\hat{Y} = H W_{\text{out}} + b_{\text{out}}.
\end{equation}

The hidden dimension is restricted to $d=16$, reducing redundancy, mitigating overfitting in data-scarce scenarios~\cite{michel2019sixteen}, and keeping the model size $\theta$ manageable for efficient tokenization and diffusion modeling.

Notably, this lightweight graph transformer serves as a shared architecture in our framework, both for collecting optimal neural parameters from source cities and for generating target-specific parameters.

%% file: Methods.tex
\section{The CDPG Framework}
\label{sec:Method}
\subsection{Problem Definition}
\label{sec:Method-Problem}
In the few-shot cross-city mobile traffic prediction task, we have abundant historical traffic data from multiple source cities, but only a few days of data for the target city. The objective is to build a strong predictor for each network element in the target city, capable of accurately predicting mobile traffic despite limited data. Specifically, we adopt a generative approach to learn neural parameters conditioned on the spatio-temporal context of each network element. Formally, let $\mathcal{C}_{S} = \{ C_1, C_2, \ldots, C_M \}$ denote the set of source cities, each with 
abundant historical traffic.  
From these source cities, we construct a dataset of optimal neural parameters:
\[
\mathcal{D}_{\theta, c} = \{ (\theta_i, c_i) \}_{i=1}^N,
\]
where $N$ is the total number of network elements in $\mathcal{C}_S$, $\theta_i \in \mathbb{R}^{d_\theta}$ denotes the optimized parameters of a lightweight Graph Transformer predictor pre-trained for network element $i$, and $c_i \in \mathbb{R}^{d_c}$ denotes the condition, which captures the spatial vector $s_i$ and temporal vector $t_i$ as $c_i = \text{concat}(s_i, t_i)$. Our goal is to learn a conditional generative function $G$ using $\mathcal{D}_{\theta, c}$, such that for any target-city network element $j$ with condition $c_j$, $G(c_j)$ produces tailored parameters $\hat{\theta}_j$. These generated parameters serve as a high-quality initialization for the target predictor, which is subsequently fine-tuned using the limited target-city traffic data to achieve optimal performance.

\subsection{Framework Overview} 
\label{sec:Method-Framework}
\begin{figure}[tb]
% \vspace{-3mm}
\centering
\includegraphics[width=0.7\linewidth]{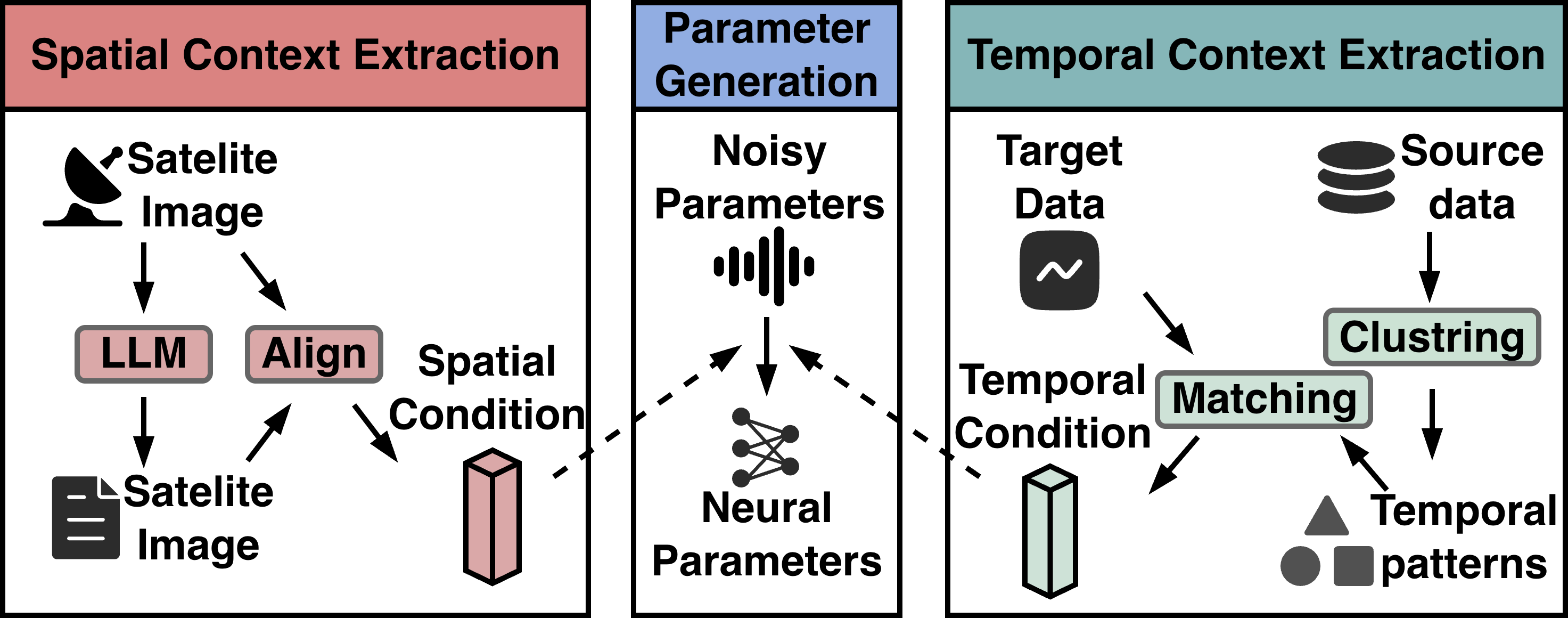}
\Description{The flowchart of the CDPG framework.}
\vspace{-2mm}
\caption{The flowchart of the CDPG framework.}
\label{fig:framework}
\vspace{-2mm}
\end{figure}

The overall framework of CDPG for the few-shot cross-city mobile traffic prediction task is illustrated in ~\figurename~\ref{fig:framework}. The framework is centered on a conditional diffusion model that generates tailored neural parameters conditioned on the target network element's context. This process employs transformer blocks for denoising and is driven by spatio-temporal conditions generated by two key components. The first component is the LLM-Enhanced Spatial Context Extraction module, which extracts spatial-functional features empowered by an LLM to form the spatial condition. The second, the Pattern-Guided Temporal Context Extraction module, constructs a taxonomy of traffic patterns via clustering from historical mobile traffic data to serve as the temporal condition. Together, these two conditions provide the rich spatio-temporal context that facilitates diffusion, enabling it to generate a specialized predictor tailored to its environment.

\subsection{Diffusion-based Neural Parameter Generation}
\label{sec:Method-Diffusion}

\begin{figure}[tb]
% \vspace{-3mm}
\centering
\includegraphics[width=0.7\linewidth]{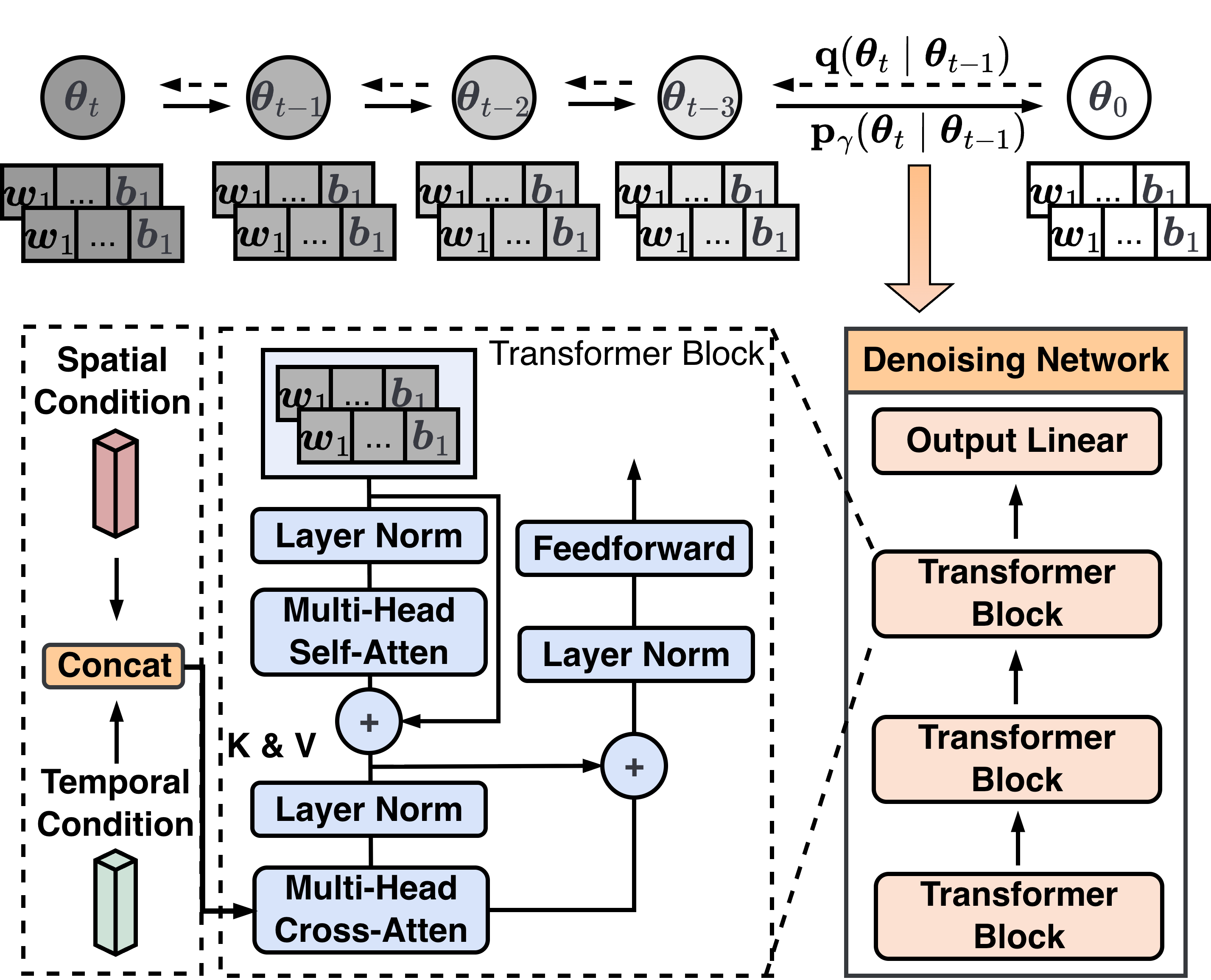}
\vspace{-2mm}
\Description{Illustration of the denoising process and the corresponding network architecture.}
\caption{Illustration of the denoising process and the corresponding network architecture.}
\label{fig:diffusion}
\vspace{-2mm}
\end{figure}

To generate specialized neural parameters tailored to the specific spatio-temporal context of each network element, we propose a conditional diffusion-based framework, as shown in~\figurename~\ref{fig:diffusion}.

\revise{ % R3-C4
The framework models the conditional distribution $p(\theta \mid c)$ and maps context $c$ to high-performing parameters $\theta$ through the reverse diffusion process.
}

\revise{ % R2-C3
Parameter-space generation provides a stability advantage in the few-shot
cross-city setting by replacing a target-distribution alignment error with a
source-informed generation error and a context error. Data-space adaptation
must estimate a source-target correspondence from the empirical target
traffic distribution. When only a few target observations are available,
sensing noise, abnormal events, and incomplete temporal coverage can make
this alignment unreliable.

CDPG instead learns an amortized mapping from abundant source-city experts to
target-specific forecasting parameters. For a target-city base station, its
adaptation process can be written as
\begin{equation}
\hat{\theta}_T=G(c_T), \qquad
\theta_T^{\mathrm{FT}}=\hat{\theta}_T+\Delta\theta_T,
\label{eq:target_parameter_adaptation}
\end{equation}
where $G(c_T)$ provides a source-informed initialization and
$\Delta\theta_T$ denotes the residual correction learned from the few-shot
target data. Under the same target-data budget, the two adaptation processes
can be conceptually decomposed as
\begin{equation}
\mathcal{E}_{\mathrm{CDPG}}
\approx
\epsilon_{\mathrm{gen}}+
\epsilon_{\mathrm{context}}+
\epsilon_{\mathrm{FT}},
\qquad
\mathcal{E}_{\mathrm{Data}}
\approx
\epsilon_{\mathrm{align}}+
\epsilon_{\mathrm{FT}},
\label{eq:adaptation_error_decomposition}
\end{equation}
where $\epsilon_{\mathrm{gen}}$ is the source-informed generation error,
$\epsilon_{\mathrm{context}}$ is the context perturbation error,
$\epsilon_{\mathrm{align}}$ is the target-distribution alignment error, and
$\epsilon_{\mathrm{FT}}$ is the residual fine-tuning error. Therefore, CDPG
is expected to be more stable when
\[
\epsilon_{\mathrm{gen}}+\epsilon_{\mathrm{context}}
<
\epsilon_{\mathrm{align}}.
\]
This condition is plausible in our setting because $\epsilon_{\mathrm{align}}$
is estimated from only a few noisy target observations, whereas
$\epsilon_{\mathrm{gen}}$ is learned offline from abundant source-city data
and $\epsilon_{\mathrm{context}}$ is partly determined by
observation-independent spatial information. Since both approaches use the
limited target data for fine-tuning, the main difference is the quality of
the initialization and the amount of target-side estimation required.

Parameter-space non-identifiability provides a further reason to use
conditional diffusion. Multiple parameter configurations may instantiate
similarly effective forecasting functions. A deterministic generator may
output a single conditional estimate that averages incompatible parameter
modes, whereas conditional diffusion models $p(\theta \mid c)$ and generates
a plausible parameter solution through iterative denoising. Therefore, the
motivation for CDPG is not that parameter-space learning is universally
superior, but that source-informed generation can replace a potentially
unstable target-distribution alignment step and provide a more informative
initialization under the few-shot cross-city conditions studied here.
}

\textbf{Collection of Optimal Neural Parameters.}
To enable parameter-space learning, we first need to collect optimal neural parameter configurations in source cities to construct a dataset. We utilize the source cities' abundant data to train a lightweight Graph Transformer predictor for each network element $i$. 
Each predictor is pre-trained independently via standard gradient descent until convergence, ensuring it becomes an "expert" tailored to its specific environment.
We collect the complete set of parameters from these experts, denoted as $\mathcal{D}_{\theta} = \{ \theta_1, \theta_2, \ldots, \theta_N \}$. This collection represents samples from the underlying optimal parameter distribution that our diffusion model aims to approximate. 
\revise{ % R1-C3
Neural parameter spaces may contain symmetries and non-identifiability, where different parameter vectors can represent similar forecasting functions. CDPG mitigates this issue through a unified expert-construction protocol. All source experts share the same lightweight Graph Transformer architecture, initialization scheme, optimizer, training schedule, and fixed layer-wise serialization order. This protocol keeps the collected parameters in a consistent coordinate system and reduces unnecessary weight-space misalignment across independently trained experts. Moreover, CDPG evaluates generated parameters by the forecasting function they instantiate, rather than by parameter-space proximity alone. The generated parameters are further fine-tuned on the few-shot target data, which corrects residual mismatch after generation.
}

\textbf{Environment Adaptive Diffusion model.}
We employ a conditional diffusion framework implemented using a Diffusion Transformer (DiT)~\cite{peebles2023scalable}. Unlike standard U-Net-based diffusion models commonly used for image data, DiT treats data as sequences, making it inherently suitable for modeling the complex dependencies within neural network parameters. 

Standard neural networks possess heterogeneous layer structures with weight matrices of varying shapes, which cannot be directly processed by a Transformer. To bridge this gap, we convert the parameter set $\mathcal{D}_{\theta}$ into a uniform sequence of tokens, adopting the strategy proposed in \cite{yuan2024spatio}. Specifically, we flatten the parameters of each layer $m$ into a 1D vector. To unify dimensions while preserving structural integrity, we calculate the Greatest Common Divisor (GCD) of parameter counts across all layers to determine a standard patch size $g$. Each flattened layer is then sliced into $n_m = N_m / g$ segments, which are linearly projected into a latent dimension $D$. This process transforms the irregular parameter set into a uniform token sequence, enabling the diffusion model to process neural parameters coherently.

The core of our approach is a context-conditioning mechanism embedded in the DiT backbone that injects spatio-temporal features from the condition into every denoising step, enabling accurate parameter generation for target-city predictors. The model is trained to reverse the forward noise injection process, effectively learning to denoise the token sequence $\theta_k$ back to $\theta_{k-1}$ under specific contexts. Inside the network, each transformer block orchestrates two critical streams of information. First, it applies Multi-Head Self-Attention to the noisy tokens to capture the intra-layer and inter-layer dependencies. Second, to enforce environmental adaptability, we inject the condition $c$ (comprising the spatial vector $s$ and the temporal vector $t$) directly via cross-attention. At each layer, the parameter tokens act as queries ($Q = \theta_k W_Q$), while the context $c$ is projected into keys ($K = c W_K$) and values ($V = c W_V$). The cross-attention operation is computed as:
\begin{equation} 
\text{CrossAttn}(\theta_k, c) = \operatorname{softmax}\left( \frac{Q K^\top}{\sqrt{d_k}} \right) V,
\label{eq:cross_attn} 
\end{equation} 
where $d_k$ is the dimensionality of the heads, and $W_Q, W_K, W_V$ are learnable matrices. Through this mechanism, the denoising network learns the conditional distribution $p(\theta \mid c)$ by optimizing the noise-prediction objective defined in (3), effectively steering the generation trajectory toward parameters optimal for the specific base-station context.

\subsection{LLM-Enhanced Spatial Context Extraction}
\label{sec:Method-LLM}
\begin{figure}[tb]
% \vspace{-3mm}
\centering
\includegraphics[width=0.63\linewidth]{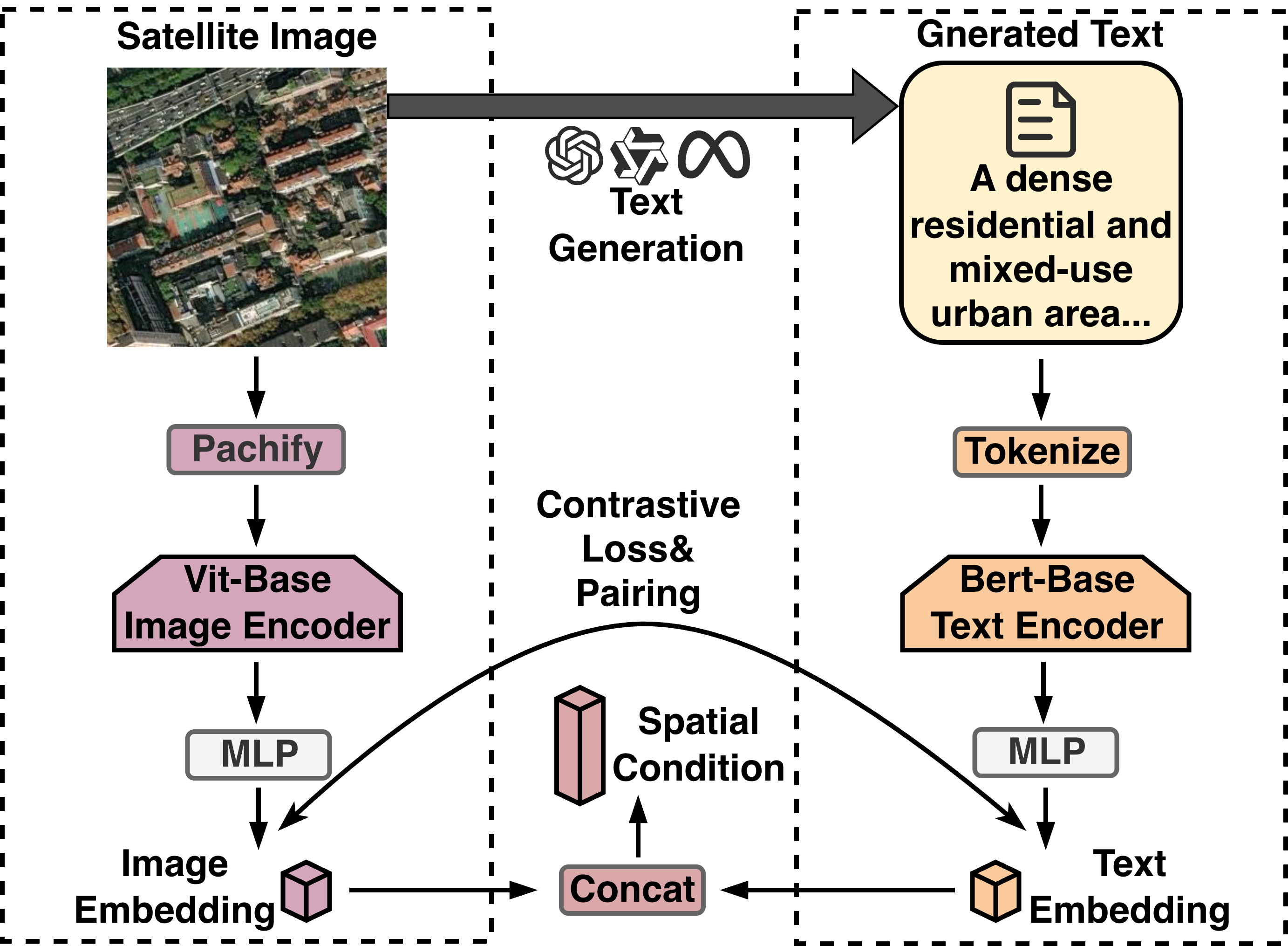}
\vspace{-2mm}
\Description{The workflow of the proposed LLM-Enhanced Spatial Context Extraction mechanism.}
\caption{The workflow of the proposed LLM-Enhanced Spatial Context Extraction mechanism.}
\label{fig:spatial}
\vspace{-3mm}
\end{figure}

To capture latent spatial dependencies invisible in raw traffic records, we propose a multi-modal spatial context extraction mechanism, as illustrated in ~\figurename~\ref{fig:spatial}. We introduce a bidirectional contrastive learning framework that aligns visual features from satellite imagery with high-level semantic descriptions generated by LLMs~\cite{liu2026llmapp, huang2025smaug}, thereby projecting them into a unified semantic space. The resulting joint representation serves as a comprehensive spatial condition $s$ for the neural parameter generation.

\textbf{Collection of Text Descriptions.}
\revise{ % R1-C4, R2-C4, R3-C2
We use pre-trained vision-language models (e.g., GPT-4o and Llama-3.2-Vision)~\cite{hurst2024gpt, dubey2024llama} to convert satellite images into functional semantic descriptions. To reduce hallucination and irrelevant visual details, we adopt a fixed structured prompt that restricts the description to observable urban cues. The prompt requires a concise 3--4 sentence description covering three aspects: $i)$ primary land use, such as residential, commercial, industrial, or mixed-use areas; $ii)$ key infrastructure, such as roads, buildings, and transportation facilities; and $iii)$ environmental features, such as green spaces and water bodies. The model is also instructed to use direct factual language and avoid unsupported assumptions.

The generated description of each satellite image is stored as a base-station-level text record. During training and inference, CDPG uses only these stored descriptions, which are encoded by the text encoder and incorporated into the spatial condition. Thus, no LLM call is required during diffusion training, parameter sampling, target fine-tuning, or online prediction.
}

\textbf{Image-text Bidirectional Contrastive Learning.}
To fuse these two modalities into a robust spatial condition, we employ a dual-encoder architecture inspired by CLIP~\cite{radford2021learning, jia2021scaling, li2021align}. We utilize a ViT-Base Encoder~\cite{dosovitskiy2020image} to encode the satellite image and a BERT-Base Encoder~\cite{devlin2019bert} to encode the corresponding textual description. 
The core objective is to learn a shared latent space in which physically distinct but functionally similar regions are mapped closely together.
After encoding, both feature vectors are projected into a shared latent space (e.g., 64 dimensions). We apply a bidirectional InfoNCE loss during training to maximize the mutual information between matched image-text pairs while minimizing it for unmatched ones.

Let $S^{ab}_{ij}$ denote the scaled similarity score between the $i$-th sample in modality $a$ and the $j$-th sample in modality $b$, defined as $S^{ab}_{ij} = \frac{\mathrm{sim}(v^{a}_i, v^{b}_j)}{\tau}$, where $\mathrm{sim}(\cdot,\cdot)$ is the cosine similarity and $\tau$ is a learnable temperature parameter. The alignment process is formulated as:
\begin{equation}
L_{\mathrm{img} \rightarrow \mathrm{text}}
= - \frac{1}{N} \sum_{i=1}^{N}
    \log \frac{\exp\left( S^{\mathrm{img,text}}_{ii} \right)}
              {\sum_{j=1}^{N} \exp\left( S^{\mathrm{img,text}}_{ij} \right)},
\label{eq:loss_img2text}
\end{equation}
\begin{equation}
L_{\mathrm{text} \rightarrow \mathrm{img}}
= - \frac{1}{N} \sum_{i=1}^{N}
    \log \frac{\exp\left( S^{\mathrm{text,img}}_{ii} \right)}
              {\sum_{j=1}^{N} \exp\left( S^{\mathrm{text,img}}_{ij} \right)}.
\label{eq:loss_text2img}
\end{equation}
The total objective is a weighted bidirectional loss, ensuring equal importance for both modalities:
\begin{equation}
L_{\mathrm{bi}}
= \alpha \, L_{\mathrm{img} \rightarrow \mathrm{text}}
+ (1-\alpha) \, L_{\mathrm{text} \rightarrow \mathrm{img}},
\label{eq:loss_bi}
\end{equation}
where $\alpha$ is set to $0.5$. Once the encoders are trained, we extract the image embedding $v^{\text{img}}$ and the text embedding $v^{\text{text}}$ for each network element. These are concatenated to form the final spatial condition vector $s = \text{concat}(v^{\text{img}}, v^{\text{text}})$, which encapsulates both physical layout and functional semantics.

\subsection{Pattern-Guided Temporal Context Extraction}
\label{sec:Method-Temporal}
\begin{figure}[tb]
% \vspace{-3mm}
\centering
\includegraphics[width=0.6\linewidth]{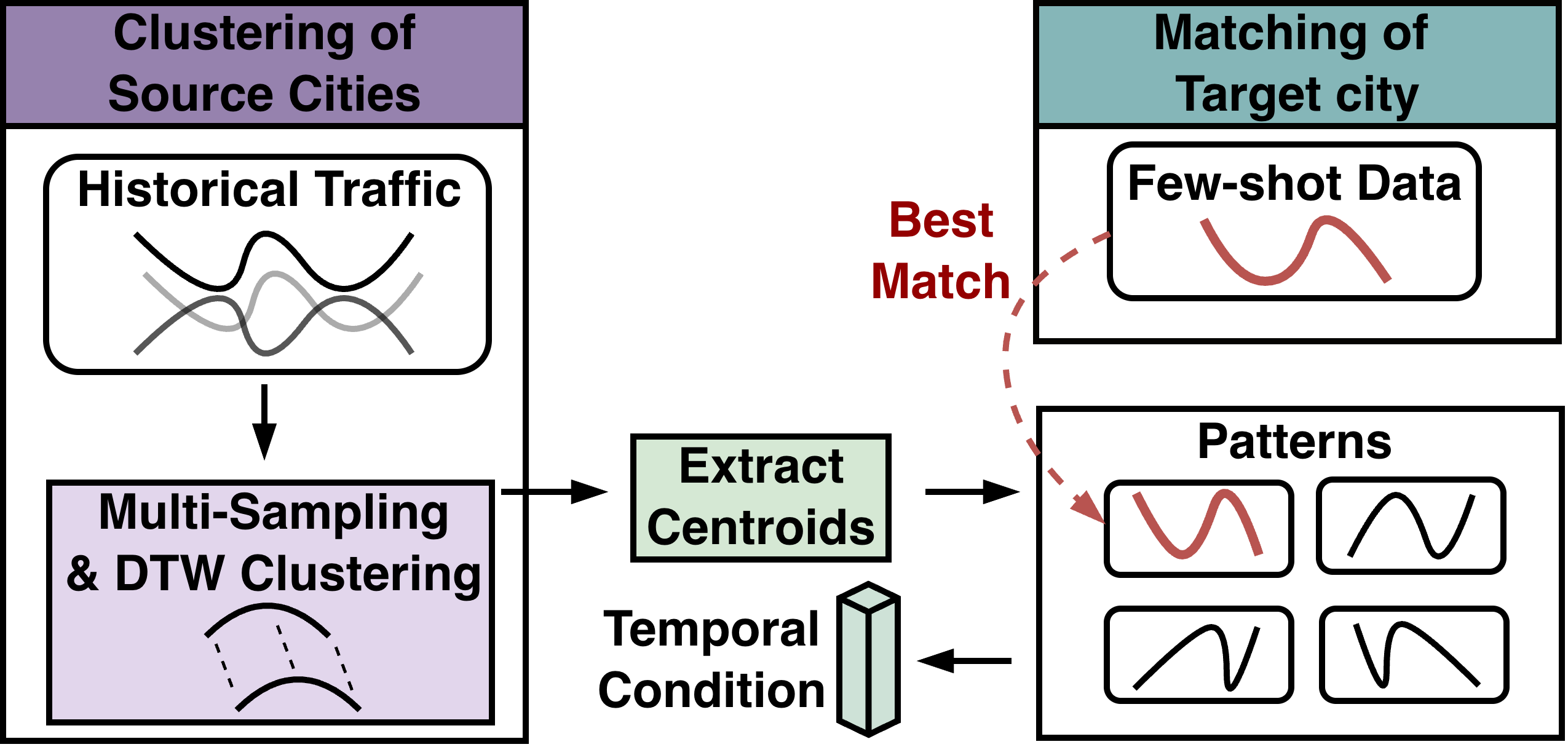}
\vspace{-2mm}
\Description{The workflow of Pattern-Guided Temporal Context Extraction.}
\caption{The workflow of Pattern-Guided Temporal Context Extraction.}
\label{fig:temporal}
\vspace{-2mm}
\end{figure}

To explicitly inform the parameter generation process with diverse traffic dynamics, we construct a robust temporal pattern taxonomy as condition vector $t$. We propose a multi-stage extraction pipeline, as shown in ~\figurename~\ref{fig:temporal}, which discovers canonical patterns from source cities and aligns target network elements with this taxonomy, thereby providing a comprehensive temporal signature.

\textbf{Mitigating Bias via Multi-Sampling.}
To mitigate the bias caused by abnormal events or seasonal fluctuations in a single time window, we treat each source network element as a multi-instance entity. Instead of relying on a single observation, we randomly sample $M$ distinct segments (i.e., $M=10$) of 128 time steps (approximately 2.6 days) from the station's entire history. This Multi-Sampling Strategy ensures that the extracted patterns reflect the station's dominant temporal behavior rather than transient noise, spanning weekdays, weekends, and holidays.

\textbf{Pattern Discovery via DTW-based Clustering.}
Standard Euclidean distance is sensitive to temporal misalignments (e.g., a morning peak occurring at 8:00 AM vs. 9:00 AM). To learn shape-invariant patterns, we cluster all sampled segments using K-means with Dynamic Time Warping (DTW) distance~\cite{petitjean2011global}. DTW allows for non-linear mapping of time axes, effectively aligning sequences with similar shapes but phase shifts.
Formally, we seek to partition the segment set $S$ into $K$ clusters to minimize the intra-cluster DTW variance:
\begin{equation}
(S^*, C^*) = \underset{S, C}{\operatorname{arg\,min}} \sum_{k=1}^{K} \sum_{x_j \in S_k} d_{\text{DTW}}(x_j, c_k),
\label{eq:clustering_objective}
\end{equation}
where $C^* = \{c_1^*, \ldots, c_K^*\}$ denotes the set of optimal centroids. Each centroid $c_k$ is updated via DTW Barycenter Averaging (DBA)~\cite{petitjean2011global}, ensuring the pattern accurately represents the mean shape of the cluster.

\textbf{Station-Level Consensus via Majority Voting.}
We sampled $M$ segments for each network element $i$, then obtain a set of cluster labels $\{l_i^{(1)}, \ldots, l_i^{(M)}\}$. To determine a robust temporal identity for the network element, we apply a Majority Voting scheme~\cite{dietterich2000ensemble}. This step filters out outlier patterns (e.g., a segment containing a rare event) and selects the most representative pattern:
\begin{equation}
l_i^* = \underset{k \in \{1, \ldots, K\}}{\operatorname{arg\,max}} \sum_{m=1}^{M} \mathbb{I}(l_i^{(m)} = k), 
\quad t_i = c_{l_i^*},
\label{eq:majority_vote}
\end{equation}
where $\mathbb{I}(\cdot)$ is the indicator function. The resulting centroid $t_i$ serves as the robust temporal condition for training.

\textbf{Few-Shot Temporal Alignment for Target Cities.} Leveraging the source traffic pattern taxonomy $C^*=\{c_1,\dots,c_K\}$ derived from the clustering process, we construct the temporal condition for target city network elements through a matching mechanism.
Under the few-shot constraint (e.g., only three days of available data), we extract a short observed traffic sequence $x_{\text{target}}$ and align it with the most semantically similar source pattern. Specifically, we compute the DTW distance between $x_{\text{target}}$ and each source pattern $c_k \in C^*$. The temporal condition vector $t_{\text{target}}$ is instantiated as the nearest pattern:
\begin{equation}
t_{\text{target}} = \underset{c_k \in C^*}{\operatorname{argmin}}  d_{\text{DTW}}(x_{\text{target}}, c_k),
\label{eq:target_matching}
\end{equation}
where $x_{\text{target}}$ denotes the observed traffic segment.
This retrieval-based alignment anchors target-domain stations to the robust source taxonomy, ensuring consistent temporal semantics for parameter generation.

\subsection{Neural Parameter Generation via Sampling}
\label{sec:Method-Sampling}
We define the condition $c$ by concatenating the spatial vector $s$ and the temporal vector $t$, which encapsulate the full spatio-temporal context of a network element. To generate specialized neural parameters for the target network element, we provide $c$ to our trained diffusion model $\epsilon_\gamma$ and initiate the reverse diffusion process in (\ref{eq:prelim_sampling}). Starting with Gaussian noise $\theta_T \sim \mathcal{N}(0, I)$, the model iteratively denoises the parameters for $T$ steps. At each step $k$, the network predicts the noise $\hat{\epsilon} = \epsilon_\gamma(\theta_k, c, k)$ and updates the parameters, yielding the final result $\theta_0$.

After sampling, the generated neural parameters $\theta_0$ serve as the initial state for the expert predictor, which is then fine-tuned using a small amount of target city data to optimize performance on the few-shot prediction task.

%% file: Experiments.tex
\section{Experiment}
\label{sec:Experiment}
In this section, we conduct extensive experiments on real-world mobile traffic datasets to evaluate CDPG's performance. We first introduce the experimental setup, covering dataset details, evaluation metrics, and baseline methods. 
\revise{ %R3-C5
Next, we report overall performance comparisons on few-shot and multi-step prediction tasks. 
}
Furthermore, we conduct detailed ablation studies and sensitivity analyses to validate the effectiveness of specific components and hyperparameter choices. Finally, we provide an analysis of computational efficiency and model complexity.

\subsection{Experiment Settings}

\begin{table*}[tb]
\centering
\footnotesize
\caption{Dataset Statistics.}
\label{table:dataset-statistics}
\vspace{-2mm}
{
    \begin{tabular}{c|ccccc}
        \toprule
        \addlinespace[3pt]
        \textbf{Dataset} &
        \textbf{Duration} &
        \textbf{Covered BSs} &
        \textbf{Satellite Images} &
        \textbf{Text Descriptions} &
        \textbf{Time Interval} \\
        \cmidrule(r{1pt}){1-6}
        \textbf{Shanghai} &
        Aug. 1st--31st, 2014 &
        4,505 &
        4,505 &
        27,030 &
        30 minutes \\
        \textbf{Nanjing} &
        Feb. 2nd--Mar. 31st, 2021 &
        8,000 &
        8,000 &
        48,000 &
        30 minutes \\
        \textbf{Nanchang} &
        July. 1st--7th, 2023 &
        7,702 &
        7,702 &
        46,212 &
        30 minutes \\
        \addlinespace[2pt]
        \bottomrule
    \end{tabular}
}
\vspace{-2mm}
\end{table*}

\subsubsection{Dataset Overview}
Our experiments are conducted on real-world datasets derived from large-scale mobile traffic networks. Table~\ref{table:dataset-statistics} presents a comprehensive summary of the dataset statistics. We categorize the data into two primary types, as detailed below.

\textbf{Mobile Traffic Datasets.} We use three real-world, large-scale mobile traffic datasets from three major Chinese cities. The Shanghai dataset consists of records from 4,505 base stations collected at 30-minute intervals between August 1 and August 31, 2014. The Nanjing dataset comprises data from 8,000 stations at 30-minute intervals between February 2 and March 31, 2021. The Nanchang dataset includes measurements from 7,702 stations, also at 30-minute intervals, between July 1–7, 2023.

\textbf{Spatial Context Datasets.} To construct the spatio-temporal context for our conditional model, we collected two additional modalities for each city. These are Satellite Imagery, retrieved from the ArcGIS mapping API based on each base station's geographic coordinates, and Semantic Descriptions, which were generated by state-of-the-art LLMs, including Qwen3-VL (235B and 32B)~\cite{bai2025qwen3}, GPT-5~\cite{wang2025capabilities}, GPT-4o~\cite{hurst2024gpt}, and Llama-3.2-Vision (90B and 11B)~\cite{dubey2024llama}.

To evaluate few-shot performance, we partition the datasets into source and target domains. The diffusion model is trained using data-rich source cities, whereas the target city is restricted to just three days of historical data for few-shot adaptation.

\subsubsection{Data Preprocessing and Source-Target Split}
\revise{ % R1-C4, R3-C3
For preprocessing, traffic records are sorted by timestamp and converted into sliding-window samples, where the previous 12 time steps are used to predict the next $H \in \{1,3,6\}$ steps. We apply $\log_{10}(x+1)$ transformation followed by node-level Z-score normalization. We use a leave-one-city-out protocol, selecting each city as the target in turn and using the other two as sources. Source experts are trained on the first 85\% of source-city time series. For the target city, the last 15\% is used for testing, and the few-shot fine-tuning set is drawn from the three days immediately before the test split to avoid temporal leakage.
}

\subsubsection{Implementation Details}
\revise{ % R1-C4, R2-C4, R3-C2
All experiments use the same lightweight Graph Transformer predictor with hidden dimension 16, one attention head, one graph transformer block, dropout 0.2, and DropPath 0.1. Source experts and target predictors are optimized with Adam using learning rate $10^{-3}$ and batch size 32. In the target city, generated parameters are used as initialization and fine-tuned only on the three-day few-shot set.
\minorrevise{ % R2-C1
During evaluation, all valid test samples, including those in the
final incomplete mini-batch, are retained when computing MAE, RMSE,
and $R^2$. This avoids potential evaluation bias from excluding
observations solely to maintain a fixed batch size.
}

For neural parameter generation, each lightweight predictor is serialized into a fixed-dimensional parameter vector using a fixed layer order. The diffusion generator is trained with learning rate $10^{-4}$, batch size 64, model dimension 512, six transformer blocks, and 500 diffusion steps. Unless otherwise stated, all baselines use the same source-target split, target fine-tuning protocol, and evaluation metrics.

Satellite images are collected from the ArcGIS World Imagery service using each base station's geographic coordinates. For each base station, we collect one $256 \times 256$ image at a fixed zoom level of 16 and store it using the corresponding base-station index. For LLM description generation, we use the fixed prompt with a maximum output length of 150 tokens and a temperature of 0, thereby avoiding additional LLM randomness during training and evaluation.

To account for stochasticity, all reported CDPG results are averaged over three independent end-to-end runs with different random seeds. Each run retrains all source experts and the diffusion generator, and repeats parameter sampling and target-city fine-tuning. Across all datasets and prediction horizons, the standard deviations are below 0.007 for MAE, 0.01 for RMSE, and 0.03 for $R^2$.

\minorrevise{ %R1-C1
For one-step MAE, we compare CDPG with ADAPTIVE using three matched independent end-to-end runs per method in each target city, under identical data splits, three-day target-data budgets, and test samples. We conduct two-sided paired \textit{t}-tests on the run-level MAE differences, defined as ADAPTIVE minus CDPG, and report unadjusted 95\% confidence intervals for the mean paired differences. We apply Holm correction across the three city-specific tests to control the family-wise error rate at 0.05.
}
}

\subsubsection{Evaluation Metrics}
We comprehensively evaluate the prediction accuracy using three widely adopted metrics: Mean Absolute Error (MAE), Root Mean Squared Error (RMSE), and the Coefficient of Determination ($R^2$).

\textbf{MAE}. MAE measures the average magnitude of the errors between the predicted and ground-truth values. It is defined as:
\begin{equation}
    \text{MAE} = \frac{1}{n} \sum_{i=1}^{n} |x_i - \hat{x}_i|,
\end{equation}
where $x_i$ and $\hat{x}_i$ denote the ground-truth and predicted values, respectively, and $n$ is the sample size.

\textbf{RMSE}. RMSE represents the quadratic mean of the differences between the observed and predicted values. The formulation is given by:
\begin{equation}
    \text{RMSE} = \sqrt{\frac{1}{n} \sum_{i=1}^{n} (x_i - \hat{x}_i)^2},
\end{equation}
where $x_i$ and $\hat{x}_i$ denote the ground-truth and predicted values, respectively, and $n$ is the sample size.

$\boldsymbol{R}^2$. $R^2$ indicates the proportion of the variance in the dependent variable that is predictable from the independent variables. It is calculated as:
\begin{equation}
    R^2 = 1 - \frac{\sum_{i=1}^{n} (x_i - \hat{x}_i)^2}{\sum_{i=1}^{n} (x_i - \bar{x})^2},
\end{equation}
where $x_i$ and $\hat{x}_i$ denote the ground-truth and predicted values, $\bar{x}$ represents the mean of the ground-truth values, and $n$ is the sample size.

\begin{table*}[tb]
\centering
\footnotesize 
\vspace{-2mm}
\caption{Comprehensive performance comparison of 1-step, 3-step, and 6-step prediction on three datasets. Best and second-best results are marked in \textbf{bold} and \underline{underlined}.}
\label{table:main_results_comprehensive}

\setlength{\tabcolsep}{5.0pt} 
\resizebox{\textwidth}{!}{
\begin{tabular}{c|c | ccc | ccc | ccc}
    \toprule

    \multirow{2.5}{*}{\textbf{Model}} & 
    \multirow{2.5}{*}{\textbf{Step}} & 
    \multicolumn{3}{c|}{\textbf{Shanghai}} & 
    \multicolumn{3}{c|}{\textbf{Nanjing}} & 
    \multicolumn{3}{c}{\textbf{Nanchang}} \\

    \cmidrule(lr){3-5} \cmidrule(lr){6-8} \cmidrule(lr){9-11}

    & & 
    \textbf{MAE} & \textbf{RMSE} & \textbf{R\textsuperscript{2}} & 
    \textbf{MAE} & \textbf{RMSE} & \textbf{R\textsuperscript{2}} & 
    \textbf{MAE} & \textbf{RMSE} & \textbf{R\textsuperscript{2}} \\
    \midrule

    \multirow{3}{*}{STGP} 
    & 1 & 0.2421 & 0.3183 & 0.7589 & 0.2547 & 0.3718 & 0.7335 & 0.2336 & 0.3526 & 0.7532 \\
    & 3 & 0.2684 & 0.3584 & 0.7157 & 0.2881 & 0.4185 & 0.6894 & 0.2673 & 0.3903 & 0.7127 \\
    & 6 & 0.2695 & 0.3563 & 0.6984 & 0.2913 & 0.4329 & 0.6584 & 0.2622 & 0.4154 & 0.6841 \\
    \midrule

    \multirow{3}{*}{UniST} 
    & 1 & 0.2287 & 0.3006 & 0.7843 & 0.2405 & 0.3512 & 0.7564 & 0.2207 & 0.3331 & 0.7881 \\
    & 3 & 0.2422 & 0.3233 & 0.7454 & 0.2599 & 0.3776 & 0.7109 & 0.2412 & 0.3522 & 0.7353 \\
    & 6 & 0.2412 & 0.3189 & 0.7289 & 0.2607 & 0.3874 & 0.6877 & 0.2346 & 0.3717 & 0.7031 \\
    \midrule

    \multirow{3}{*}{CCTP} 
    & 1 & 0.2144 & 0.2819 & 0.8133 & 0.2256 & 0.3293 & 0.7831 & 0.2069 & 0.3124 & 0.8166 \\
    & 3 & 0.2172 & 0.2900 & 0.7811 & 0.2331 & 0.3386 & 0.7448 & 0.2163 & 0.3158 & 0.7700 \\
    & 6 & 0.2154 & 0.2848 & 0.7601 & 0.2328 & 0.3459 & 0.7280 & 0.2095 & 0.3319 & 0.7431 \\
    \midrule

    \multirow{3}{*}{MetaST} 
    & 1 & 0.2105 & 0.2776 & 0.8169 & 0.2224 & 0.3257 & 0.7882 & 0.2026 & 0.3085 & 0.8215 \\
    & 3 & 0.2156 & 0.2884 & 0.7849 & 0.2314 & 0.3361 & 0.7518 & 0.2115 & 0.3112 & 0.7764 \\
    & 6 & 0.2132 & 0.2828 & 0.7674 & 0.2296 & 0.3420 & 0.7359 & 0.2078 & 0.3290 & 0.7508 \\
    \midrule

    \multirow{3}{*}{MGSTPAT} 
    & 1 & 0.2029 & 0.2668 & 0.8276 & 0.2134 & 0.3115 & 0.8062 & 0.1958 & 0.2955 & 0.8307 \\
    & 3 & 0.2012 & 0.2697 & 0.8012 & 0.2192 & 0.3134 & 0.7652 & 0.2003 & 0.2922 & 0.7971 \\
    & 6 & 0.2089 & 0.2762 & 0.7823 & 0.2258 & 0.3355 & 0.7466 & 0.2032 & 0.3219 & 0.7713 \\
    \midrule

    \multirow{3}{*}{ADAPTIVE} 
    & 1 & \underline{0.1949} & \underline{0.2563} & \underline{0.8430} & \underline{0.2051} & \underline{0.2994} & \underline{0.8105} & \underline{0.1882} & \underline{0.2839} & 0.8358 \\
    & 3 & 0.2068 & 0.2762 & 0.7943 & 0.2253 & 0.3209 & 0.7574 & 0.2059 & 0.3004 & 0.7828 \\
    & 6 & \underline{0.2030} & \underline{0.2674} & \underline{0.7955} & \underline{0.2197} & \underline{0.3205} & \underline{0.7629} & \underline{0.1965} & \underline{0.3153} & \underline{0.7835} \\
    \midrule

    \multirow{3}{*}{ST-WA} 
    & 1 & 0.1985 & 0.2608 & 0.8374 & 0.2096 & 0.3058 & 0.8032 & 0.1914 & 0.2887 & 0.8341 \\
    & 3 & 0.2047 & 0.2731 & 0.7985 & 0.2216 & 0.3178 & 0.7619 & 0.2018 & 0.2945 & 0.7914 \\
    & 6 & 0.2061 & 0.2712 & 0.7894 & 0.2227 & 0.3264 & 0.7568 & 0.1988 & 0.3181 & 0.7786 \\
    \midrule

    \multirow{3}{*}{SynEVO} 
    & 1 & 0.1971 & 0.2592 & 0.8390 & 0.2078 & 0.3036 & 0.8072 & 0.1903 & 0.2869 & 0.8371 \\
    & 3 & 0.2024 & 0.2718 & 0.8004 & 0.2187 & 0.3147 & 0.7688 & 0.1989 & 0.2914 & 0.8007 \\
    & 6 & 0.2044 & 0.2690 & 0.7931 & 0.2209 & 0.3228 & 0.7608 & 0.1972 & 0.3165 & 0.7819 \\
    \midrule

    \multirow{3}{*}{GPD} 
    & 1 & 0.1966 & 0.2587 & 0.8412 & 0.2072 & 0.3028 & 0.8089 & 0.1896 & 0.2861 & \underline{0.8378} \\
    & 3 & \underline{0.1958} & \underline{0.2618} & \underline{0.8230} & \underline{0.2118} & \underline{0.3062} & \underline{0.7901} & \underline{0.1887} & \underline{0.2855} & \underline{0.8341} \\
    & 6 & 0.2049 & 0.2704 & 0.7916 & 0.2215 & 0.3240 & 0.7587 & 0.1980 & 0.3172 & 0.7802 \\
    \midrule

    \multirow{3}{*}{\textbf{CDPG}} 
    & 1 & \textbf{0.1786} & \textbf{0.2445} & \textbf{0.8759} & \textbf{0.1929} & \textbf{0.2875} & \textbf{0.8519} & \textbf{0.1699} & \textbf{0.2639} & \textbf{0.8995} \\
    & 3 & \textbf{0.1875} & \textbf{0.2542} & \textbf{0.8496} & \textbf{0.2025} & \textbf{0.2990} & \textbf{0.8082} & \textbf{0.1783} & \textbf{0.2749} & \textbf{0.8663} \\
    & 6 & \textbf{0.1911} & \textbf{0.2591} & \textbf{0.8329} & \textbf{0.2050} & \textbf{0.3047} & \textbf{0.8014} & \textbf{0.1813} & \textbf{0.2871} & \textbf{0.8455} \\
    
    \midrule
    \midrule

    \multirow{3}{*}{Improvement} 
    & 1 & +8.36\% & +4.60\% & +3.90\% & +5.95\% & +3.97\% & +5.11\% & +9.72\% & +7.04\% & +7.36\% \\
    & 3 & +4.24\% & +2.90\% & +3.23\% & +4.39\% & +2.35\% & +2.29\% & +5.51\% & +3.71\% & +3.86\% \\
    & 6 & +5.86\% & +3.10\% & +4.70\% & +6.69\% & +4.93\% & +5.05\% & +7.74\% & +8.94\% & +7.91\% \\

    \bottomrule
\end{tabular}
}
\end{table*}

\begin{table*}[tb]
\centering
\footnotesize
\vspace{-2mm}
\caption{Component and adaptation ablation study for 1-step-ahead prediction.
Best and second-best results are marked in \textbf{bold} and
\underline{underlined}.
All variants include target-city fine-tuning unless marked ``w/o FT''.
}
\label{table:component_ablation_study}

\renewcommand{\arraystretch}{1.00}
\setlength{\tabcolsep}{8pt}

\resizebox{\textwidth}{!}{
\begin{tabular}{c | ccc | ccc | ccc}
    \toprule
    \multirow{2}{*}{\textbf{Model}} &
    \multicolumn{3}{c|}{\textbf{Shanghai}} &
    \multicolumn{3}{c|}{\textbf{Nanjing}} &
    \multicolumn{3}{c}{\textbf{Nanchang}} \\

    \cmidrule(lr){2-4}
    \cmidrule(lr){5-7}
    \cmidrule(lr){8-10}

    & \textbf{MAE} & \textbf{RMSE} & \textbf{R\textsuperscript{2}}
    & \textbf{MAE} & \textbf{RMSE} & \textbf{R\textsuperscript{2}}
    & \textbf{MAE} & \textbf{RMSE} & \textbf{R\textsuperscript{2}} \\
    \midrule

    w/o Condition &
    0.1963 & 0.2765 & 0.8413 &
    0.2142 & 0.3278 & 0.8075 &
    0.1895 & 0.3041 & 0.8665 \\

    w/o Temporal &
    0.1846 & \underline{0.2542} & \underline{0.8659} &
    \underline{0.1975} & 0.3024 & 0.8362 &
    \underline{0.1743} & 0.2748 & 0.8910 \\

    w/o Spatial &
    0.1891 & 0.2673 & 0.8517 &
    0.2057 & 0.3146 & 0.8227 &
    0.1792 & 0.2835 & 0.8840 \\
    \midrule

    CDPG-Image &
    \underline{0.1834} & 0.2549 & 0.8651 &
    0.1983 & \underline{0.2992} & \underline{0.8396} &
    0.1745 & \underline{0.2726} & \underline{0.8928} \\

    CDPG-LLM &
    0.1848 & 0.2601 & 0.8596 &
    0.2015 & 0.3057 & 0.8326 &
    0.1762 & 0.2743 & 0.8914 \\

    w/o Contrastive &
    0.1839 & 0.2560 & 0.8643 &
    0.1996 & 0.3010 & 0.8381 &
    0.1751 & 0.2735 & 0.8919 \\
    \midrule

    CDPG-HyperNet &
    0.2048 & 0.2896 & 0.8234 &
    0.2269 & 0.3475 & 0.7816 &
    0.1998 & 0.3224 & 0.8427 \\

    CDPG-CVAE &
    0.1868 & 0.2572 & 0.8614 &
    0.2018 & 0.3069 & 0.8351 &
    0.1765 & 0.2772 & 0.8902 \\
    \midrule

    CDPG-HyperNet w/o FT &
    0.2254 & 0.3182 & 0.7896 &
    0.2497 & 0.3781 & 0.7446 &
    0.2213 & 0.3546 & 0.8193 \\
    
    CDPG-CVAE w/o FT &
    0.2028 & 0.2845 & 0.8318 &
    0.2206 & 0.3372 & 0.7969 &
    0.1931 & 0.3068 & 0.8648 \\

    CDPG w/o FT &
    0.1947 & 0.2742 & 0.8438 &
    0.2124 & 0.3256 & 0.8106 &
    0.1878 & 0.3015 & 0.8694 \\

    Random Init. + FT &
    0.2590 & 0.3406 & 0.7058 &
    0.2725 & 0.3978 & 0.6822 &
    0.2500 & 0.3773 & 0.7005 \\
    \midrule

    \textbf{CDPG} &
    \textbf{0.1786} & \textbf{0.2445} & \textbf{0.8759} &
    \textbf{0.1929} & \textbf{0.2875} & \textbf{0.8519} &
    \textbf{0.1699} & \textbf{0.2639} & \textbf{0.8995} \\

    \bottomrule
\end{tabular}
}
\vspace{-2mm}
\end{table*}

\subsubsection{Baselines}
\revise{ % R1-C2 R3-C5
To evaluate the effectiveness of our proposed framework, we compare CDPG against nine representative baselines under the few-shot cross-city setting, covering foundation-model-based prediction, data-space transfer, meta-learning, non-diffusion parameter generation, cross-domain adaptation, and diffusion-based neural parameter generation.
}

\begin{itemize}
\item \textbf{CCTP}~\cite{wu2021deep}. CCTP is a cross-city deep transfer learning framework that pre-trains a spatial-temporal model (GASTN) on a source city. It aligns domains using a GAN-based adapter and an inter-city region mapping for knowledge transfer.

\item \textbf{ADAPTIVE}~\cite{zhang2023deep}. ADAPTIVE leverages an Urban Knowledge Graph (UKG) to align base station representations. It then uses a source-trained, feature-enhanced GAN to generate traffic for the target domain by conditioning on these aligned representations.

\item \textbf{MGSTPAT}~\cite{mo2025knowledge}. MGSTPAT acquires multi-granularity knowledge via adaptive clustering to automatically discover source patterns (without a fixed $K$). It then employs attention-based retrieval to selectively transfer the most relevant patterns to the target domain.

\item \textbf{STGP}~\cite{hu2024prompt}. STGP introduces a prompt-based paradigm that unifies diverse spatio-temporal tasks into a shared data reconstruction template. It employs a two-stage prompting pipeline (comprising domain and task prompts) to efficiently adapt frozen pre-trained models to the target domain.

\item \textbf{UniST}~\cite{yuan2024unist}. UniST is a universal framework pre-trained via masked token modeling that adapts a frozen base model to new scenarios using knowledge-guided prompts. By querying spatial and temporal memory pools, it integrates multi-view domain knowledge to generate context-aware prompts for effective adaptation.

\revise{ % R1-C2
\item \textbf{MetaST}~\cite{yao2019learning}. MetaST is a meta-learning-based spatial-temporal prediction method with a Model-Agnostic Meta-Learning (MAML)-style initialization strategy. It learns a generalized initialization from multiple source cities for fast adaptation to data-scarce target cities.

\item \textbf{ST-WA}~\cite{cirstea2022towards}. ST-WA is a spatio-temporal aware forecasting method with a non-diffusion parameter-generation strategy. It generates location-specific and time-varying forecasting parameters from stochastic variables to capture heterogeneous traffic dynamics.

\item \textbf{GPD}~\cite{yuan2024spatio}. GPD is a spatio-temporal few-shot learning method with a diffusion-based neural parameter generation strategy. It generates task-specific neural network parameters for fast adaptation to few-shot spatio-temporal prediction tasks.
}

\revise{ % R3-C5
\item \textbf{SynEVO}~\cite{liu2025synevo}. SynEVO is a cross-domain spatio-temporal adaptation framework that evolves shared and domain-specific knowledge across domains. We adapt it by treating each city as a domain and using the few-shot target-city data for adaptation.
}

\end{itemize}
All experiments were conducted on a workstation equipped with four NVIDIA GeForce RTX 4090 GPUs.

\subsection{Overall Performance Comparison}
\label{sec:exp_overall}
\revise{ % R1-C2 R3-C5
The results are presented in Table~\ref{table:main_results_comprehensive}. Across all datasets and prediction steps, CDPG achieves the best performance on MAE, RMSE, and $R^2$. On average, CDPG reduces MAE by 6.50\%, reduces RMSE by 4.62\%, and improves $R^2$ by 4.82\% compared with the second-best results, demonstrating its consistent advantage under few-shot cross-city prediction.

We include four additional task-aligned baselines to better examine the source of this advantage. MetaST represents a Model-Agnostic Meta-Learning (MAML)-style initialization strategy for fast target-city adaptation. ST-WA represents a non-diffusion parameter-generation strategy that produces location-specific and time-varying forecasting parameters. GPD is the closest diffusion-based neural parameter generation baseline, and SynEVO represents recent cross-domain spatio-temporal adaptation. 
Compared with these baselines, CDPG remains consistently superior, showing that conditional diffusion-based parameter generation with base-station-level spatio-temporal conditions is more effective in our setting.

In the 1-step prediction task, CDPG consistently outperforms all baselines. It reduces MAE by up to 9.72\% and RMSE by up to 7.04\%, and improves $R^2$ by up to 7.36\%. 
\minorrevise{ %R1-C1
To assess the statistical reliability of these MAE gains, we compare CDPG with ADAPTIVE, the strongest MAE baseline in all three target cities. The mean paired MAE reductions are 0.0163 (95\% CI: [0.0101, 0.0225]), 0.0122 (95\% CI: [0.0047, 0.0197]), and 0.0183 (95\% CI: [0.0071, 0.0295]) for Shanghai, Nanjing, and Nanchang, respectively. The corresponding Holm-adjusted $p$-values are 0.0233, 0.0391, and 0.0391, all below 0.05, indicating statistically significant MAE improvements over ADAPTIVE under the evaluated settings. The confidence intervals are pointwise and unadjusted for multiplicity.
}
For the 3-step task, where GPD becomes the strongest competing method, CDPG still reduces MAE and RMSE by 4.71\% and 2.99\% on average across the three datasets, respectively, while improving $R^2$ by 3.13\% on average. For the 6-step task, CDPG achieves up to 7.74\% MAE
reduction, 8.94\% RMSE reduction, and 7.91\% $R^2$ improvement, confirming its robustness for longer-horizon forecasting.

The comparison also clarifies the limitations of different transfer paradigms under few-shot cross-city mobile traffic prediction. ADAPTIVE and MGSTPAT remain competitive among data-space transfer methods, benefiting from knowledge-graph alignment and multi-granularity pattern transfer, respectively. MGSTPAT also serves as a strong retrieval-based transfer baseline, covering the nearest-source transfer family by selecting relevant source patterns for the target city. However, these methods still depend on matching or transferring observed traffic patterns across cities, which can be affected by sparse target observations and cross-city distribution shifts. MetaST learns a shared initialization for fast target-city adaptation, but a single initialization is less expressive for heterogeneous base-station-level dynamics. ST-WA generates location-specific and time-varying forecasting parameters, but it does not explicitly model the conditional distribution of full predictor parameters under cross-city few-shot transfer. SynEVO adapts shared cross-domain knowledge, but its adaptation is mainly performed at the domain/model level and may not capture fine-grained base-station heterogeneity. GPD performs diffusive neural generation for general spatio-temporal few-shot learning, but it relies on coarser domain-level conditions and does not incorporate multimodal spatial semantics or explicit temporal pattern taxonomy. In contrast, CDPG generates lightweight predictor parameters conditioned on multimodal spatial semantics and temporal traffic patterns, which better captures base-station-level heterogeneity under cross-city distribution shifts.
}

\subsection{Ablation Studies}
We conduct ablation studies on the 1-step-ahead prediction task, using the previous 12 time steps as input, to validate the necessity of each key design in CDPG. Specifically, we compare CDPG with twelve variants that remove or replace different components:
\begin{itemize}
  
\item \textbf{w/o Condition.} The entire spatio-temporal condition is removed, and parameters are generated unconditionally with diffusion model.

\item \textbf{w/o Temporal.} The temporal condition is omitted; only the spatial condition is used as the parameter generation condition.

\item \textbf{w/o Spatial.} The temporal condition is retained, while the spatial condition is excluded from the parameter generation condition.

\item \textbf{CDPG-Image.} The spatial condition is provided solely by satellite imagery embeddings, without LLM-generated semantic descriptions.

\item \textbf{CDPG-LLM.} The spatial condition is derived exclusively from textual descriptions, excluding satellite imagery embeddings.

\revise{ % R2-C2
\item \textbf{w/o Contrastive.} The bidirectional image-text contrastive learning objective is removed, and the image and text encoder embeddings are directly concatenated.
}

\revise{ % R1-C2、R2-C2
\item \textbf{CDPG-HyperNet.} The diffusion generator is replaced with an MLP-based hypernetwork that directly maps the spatio-temporal condition to predictor parameters.

\item \textbf{CDPG-CVAE.} The diffusion generator is replaced with a conditional variational autoencoder for parameter generation.
}

\revise{ % R1-C3
\minorrevise{ %R1-C2
\item \textbf{CDPG-HyperNet w/o FT.}
Parameters generated by the HyperNetwork are directly evaluated on
the target city without fine-tuning.

\item \textbf{CDPG-CVAE w/o FT.}
Parameters generated by the CVAE are directly evaluated on the target
city without fine-tuning.

\item \textbf{CDPG w/o FT.}
Parameters generated by diffusion are directly evaluated on the target
city without fine-tuning.
}

\item \textbf{Random Init. + FT.} The predictor is randomly initialized and fine-tuned (FT) using the same target-city data and optimization settings as CDPG.
}
\end{itemize}

\revise{ % R1-C2、R2-C2
Table~\ref{table:component_ablation_study} reports the results across the three datasets. CDPG consistently achieves the best performance, confirming that its gains come from the joint effect of conditional diffusion-based parameter generation, multimodal spatial modeling, and temporal pattern guidance.

\minorrevise{ % R2-W3
Removing the spatio-temporal condition degrades performance.
Among the individual conditions, removing spatial context causes
a larger loss than removing temporal context. In w/o Spatial,
both image- and text-derived spatial conditions are removed,
while temporal conditioning and the diffusion-based generation
and fine-tuning procedures are retained. Relative to this variant,
the full model reduces MAE by 5.55\%, 6.22\%, and 5.19\% in Shanghai,
Nanjing, and Nanchang, respectively, averaging 5.66\% across cities.
These results support the contribution of multimodal spatial
information to CDPG's performance.
}

Both CDPG-Image and CDPG-LLM underperform the full model, demonstrating the complementarity between visual and textual spatial information. Satellite imagery captures physical layout and land-use patterns, while LLM-generated descriptions provide higher-level functional semantics. The w/o Contrastive variant performs close to the single-modality variants and worse than CDPG, suggesting that simple feature concatenation is insufficient. The contrastive objective aligns the two modalities into a more coherent spatial condition for parameter generation.

\minorrevise{ % R1-C2, R2-W3
The MLP-based HyperNetwork used here directly maps each condition
to a single parameter vector. CVAE introduces stochastic generation
through latent-variable reconstruction and KL regularization,
whereas diffusion learns conditional denoising across multiple
noise levels and generates parameters iteratively. All three
variants use the same source-expert parameter dataset, spatial
and temporal conditioning inputs, predictor architecture, and
target-city fine-tuning protocol.
Table~\ref{table:component_ablation_study} compares the generated
predictors before and after fine-tuning. Before fine-tuning,
the MAEs in Shanghai, Nanjing, and Nanchang are 0.2254, 0.2497,
and 0.2213 for HyperNet; 0.2028, 0.2206, and 0.1931 for CVAE;
and 0.1947, 0.2124, and 0.1878 for diffusion, respectively.
Averaging the city-specific relative reductions, diffusion lowers
MAE by 14.57\% compared with HyperNet and 3.49\% compared with CVAE.
After fine-tuning, its corresponding advantages remain 14.25\%
and 4.18\%. These comparisons support the benefit of diffusion-based
generator design under identical conditioning information, with
its advantage present before fine-tuning and retained afterward.
Together with the spatial ablation, they show that generator
design and multimodal spatial context jointly contribute to
CDPG's performance.
}
}
\revise{ % R1-C3
The adaptation ablations in Table 3 separate the contribution of generated parameters from that of target-city fine-tuning. Even without target-city fine-tuning, the CDPG w/o FT variant substantially outperforms Random Init. + FT
across all three datasets. On average, it reduces MAE and RMSE by 23.9\% and 19.2\%, respectively, while achieving an absolute $R^2$ gain of 0.145. This result shows that the generated parameters already encode transferable forecasting functions, despite receiving no target-city parameter updates.

Fine-tuning from the generated parameters provides further improvements. Compared with CDPG w/o FT, the full CDPG reduces MAE and RMSE by an additional 9.0\% and 11.7\% on average, respectively, and yields an absolute $R^2$ gain
of 0.035. Together, these comparisons show that CDPG's gains cannot be attributed to target-city fine-tuning alone. The generated initialization provides effective cross-city knowledge transfer, while fine-tuning further adapts the predictor to the remaining target-city mismatch.
}

\begin{table*}[tb]
\centering
\footnotesize 
% \vspace{-2mm}
\caption{Performance comparison of LLM choice for 1-step-ahead prediction across three datasets. Best and second-best results are marked in \textbf{bold} and \underline{underlined}.}
\label{table:llm_choice}

\renewcommand{\arraystretch}{1.0} 

\setlength{\tabcolsep}{8pt}
\resizebox{\textwidth}{!}{
\begin{tabular}{c | ccc | ccc | ccc}
    \toprule

    \multirow{2.5}{*}{\textbf{LLM}} & 
    \multicolumn{3}{c|}{\textbf{Shanghai}} & 
    \multicolumn{3}{c|}{\textbf{Nanjing}} & 
    \multicolumn{3}{c}{\textbf{Nanchang}} \\

    \cmidrule(lr){2-4} \cmidrule(lr){5-7} \cmidrule(lr){8-10}

    & \textbf{MAE} & \textbf{RMSE} & \textbf{R\textsuperscript{2}} & 
      \textbf{MAE} & \textbf{RMSE} & \textbf{R\textsuperscript{2}} & 
      \textbf{MAE} & \textbf{RMSE} & \textbf{R\textsuperscript{2}} \\
    \midrule
    
    Llama-3.2-Vision (11B) & 
    0.1904 & 0.2695 & 0.8492 & 
    0.2178 & 0.3274 & 0.8079 & 
    0.1812 & 0.2835 & 0.8840 \\
    
    Qwen3-VL (32B) & 
    0.1856 & 0.2563 & 0.8636 & 
    0.2035 & 0.3027 & 0.8358 & 
    0.1748 & 0.2742 & 0.8915 \\
    
    Llama-3.2-Vision (90B) & 
    0.1824 & 0.2528 & 0.8673 & 
    0.2013 & 0.2976 & 0.8413 & 
    0.1785 & 0.2764 & 0.8898 \\
    
    GPT-4o & 
    0.1815 & 0.2522 & 0.8680 & 
    0.1974 & 0.2985 & 0.8404 & 
    0.1769 & 0.2813 & 0.8858 \\
    
    GPT-5 & 
    \underline{0.1793} & \underline{0.2486} & \underline{0.8717} & 
    \underline{0.1962} & \underline{0.2958} & \underline{0.8432} & 
    \underline{0.1715} & \underline{0.2716} & \underline{0.8935} \\
    
    \textbf{Qwen3-VL (235B)} & 
    \textbf{0.1786} & \textbf{0.2445} & \textbf{0.8759} & 
    \textbf{0.1929} & \textbf{0.2875} & \textbf{0.8519} & 
    \textbf{0.1699} & \textbf{0.2639} & \textbf{0.8995} \\
    
    \bottomrule
\end{tabular}
}
\vspace{-2mm}
\end{table*}

\begin{figure*}[tb] 
\centering
\subfloat[Temporal Clusters]
{\includegraphics[width=0.24\textwidth]{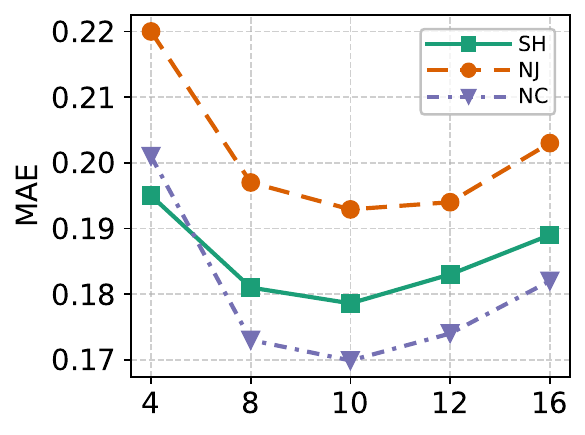}
\label{fig:cluster_k}}
\hfil
\subfloat[Model Dimension]{\includegraphics[width=0.24\textwidth]{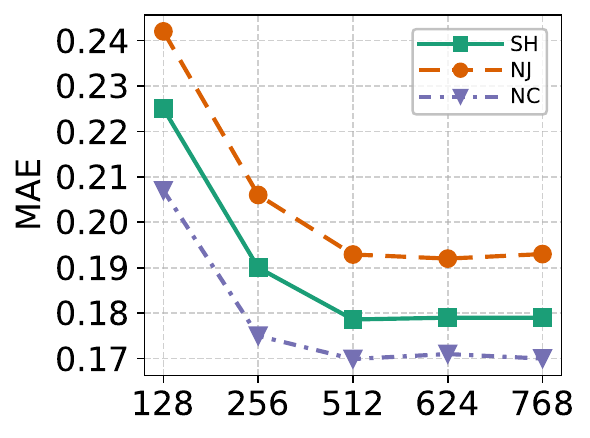}
\label{fig:hyper_dim}}
\hfil 
\subfloat[Model Depth]{\includegraphics[width=0.24\textwidth]{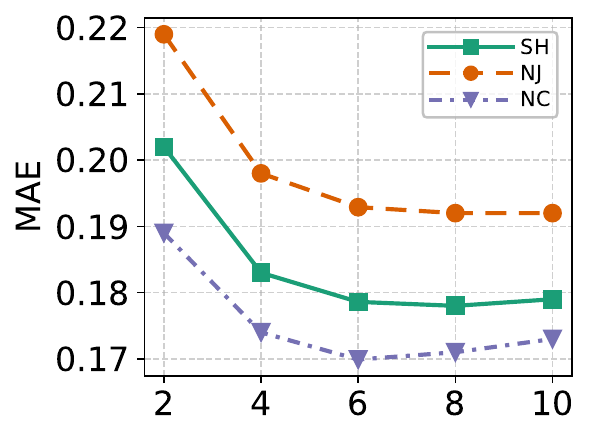}
\label{fig:hyper_layers}}
\hfil 
\subfloat[Diffusion Steps]{\includegraphics[width=0.24\textwidth]{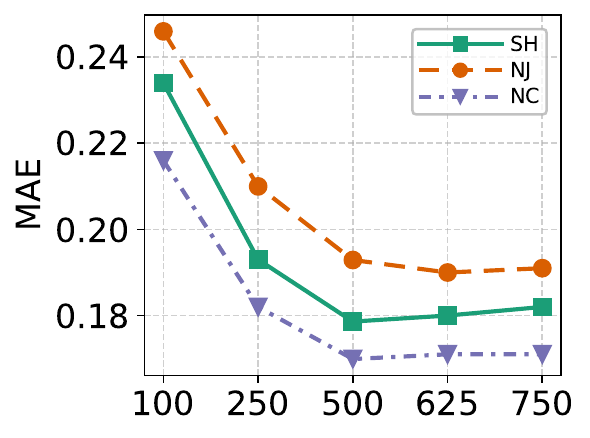}
\label{fig:hyper_steps}}
\vspace{-3mm}
\Description{Sensitivity analysis of key hyperparameters on MAE across three datasets (Task: 1-step-ahead prediction): (a) Number of temporal clusters, (b) Model dimension, (c) Model depth, and (d) Diffusion steps.}
\caption{Sensitivity analysis of key hyperparameters on MAE across three datasets (Task: 1-step-ahead prediction): (a) Number of temporal clusters, (b) Model dimension, (c) Model depth, and (d) Diffusion steps.}
\label{fig:hyper_diffusion} 
\end{figure*}

\subsection{Analysis of Hyperparameter and Sensitivity}
We verify the robustness of CDPG by conducting a series of analyses regarding data configurations and key hyperparameters on the 1-step-ahead prediction task.

\begin{table}[tb]
\centering
\footnotesize
\captionsetup{width=1.0\columnwidth}
\caption{Impact of source city diversity on 1-step-ahead prediction performance. SH: Shanghai; NJ: Nanjing; NC: Nanchang. Best results are marked in \textbf{bold}.}
\label{table:source_city_config}

\renewcommand{\arraystretch}{1.0}
\setlength{\tabcolsep}{8pt}

\begin{tabular}{c|c|c|c|c}
    \toprule
    \textbf{Target} & \textbf{Source} & \textbf{MAE} & \textbf{RMSE} & \textbf{R\textsuperscript{2}} \\
    \midrule

    \multirow{3}{*}{SH} 
    & NC       & 0.1904 & 0.2685 & 0.8503 \\
    & NJ       & 0.1882 & 0.2617 & 0.8578 \\
    & \textbf{NC + NJ} & \textbf{0.1786} & \textbf{0.2445} & \textbf{0.8759} \\
    \midrule

    \multirow{3}{*}{NJ} 
    & SH       & 0.2035 & 0.3152 & 0.8220 \\
    & NC       & 0.2051 & 0.3108 & 0.8269 \\
    & \textbf{SH + NC} & \textbf{0.1929} & \textbf{0.2875} & \textbf{0.8519} \\
    \midrule

    \multirow{3}{*}{NC} 
    & SH       & 0.1803 & 0.2780 & 0.8885 \\ 
    & NJ       & 0.1781 & 0.2755 & 0.8905 \\ 
    & \textbf{SH + NJ} & \textbf{0.1699} & \textbf{0.2639} & \textbf{0.8995} \\

    \bottomrule
\end{tabular}
\vspace{-2mm}
\end{table}

\subsubsection{Effect of LLM Selection in Spatial Context Modeling}
The quality of the text descriptions directly influences the semantic richness of the spatial condition $s$. We compare the performance of CDPG when text descriptions are generated by several LLMs at various scales: Qwen3-VL (235B and 32B), GPT-5, GPT-4o, and Llama-3.2-Vision (90B and 11B). The experimental results, summarized in Table~\ref{table:llm_choice}, reveal a distinct correlation between LLM capability and downstream prediction accuracy. Specifically, the Qwen3-VL (235B) model achieves superior performance, thanks to its robust vision-language alignment, which accurately extracts high-level functional semantics from satellite imagery. In contrast, Llama-3.2-Vision (11B) suffers a degradation of up to 13.0\% in MAE compared to the best model. We attribute this substantial gap to the limited reasoning capabilities of smaller models, which tend to produce generic descriptions or hallucinations that mislead the parameter generation process. The performance of intermediate models, including GPT-5, GPT-4o, and Llama-3.2-Vision (90B), falls between these two extremes, confirming the significant impact of model scale.

\subsubsection{Influence of Source-city Configurations}
We investigate how the diversity of source cities affects knowledge transfer performance on the 1-step-ahead prediction task. To illustrate, we use Nanchang as the target city and evaluate three source configurations: $i)$ using only Shanghai, $ii)$ using only Nanjing, and $iii)$ using both Shanghai and Nanjing. As shown in Table~\ref{table:source_city_config}, utilizing a combination of multiple source cities consistently yields lower errors compared to using any single source city. For instance, when targeting Nanchang, the dual-source setting (Shanghai + Nanjing) reduces MAE by approximately 4-6\% and RMSE by approximately 4-5\% compared to the single-source baselines. This confirms that the diffusion model can effectively distill useful and transferable knowledge from diverse urban environments across source cities, thereby improving its adaptability to the unseen target city.

\subsubsection{Parameter Sensitivity}
\revise{ % R2-C2
We analyze four key hyperparameters of CDPG: the number of temporal clusters, DiT model dimension, DiT depth, and diffusion steps, as shown in \figurename~\ref{fig:hyper_diffusion}. These parameters correspond to temporal pattern granularity, denoising backbone capacity, and generation refinement strength.

For temporal clustering, performance improves as the number of clusters increases, because finer clusters capture more diverse traffic patterns. The gain saturates after 10 clusters and slightly decreases with more clusters, suggesting that overly fine-grained temporal taxonomies may introduce noisy or source-specific patterns. For the DiT backbone, increasing model dimension and depth improves performance at first, but the benefit saturates at dimension 512 and depth 6. This indicates that DiT needs enough capacity to model dependencies among serialized parameter tokens, while excessive capacity brings limited gains and higher cost. For diffusion steps, reducing the number from 500 to 100 causes about 30\% performance degradation, confirming the importance of iterative denoising. Increasing beyond 500 brings marginal accuracy gains but nearly proportional inference latency.

Based on these observations, we set the number of temporal clusters to 10, DiT model dimension to 512, DiT depth to 6, and diffusion steps to 500 across all datasets.
}

\subsection{Model Complexity and Deployment Efficiency}
\label{sec:efficiency_analysis}
\begin{figure}[tb]
% \vspace{-3mm}
\centering
\includegraphics[width=0.72\linewidth]{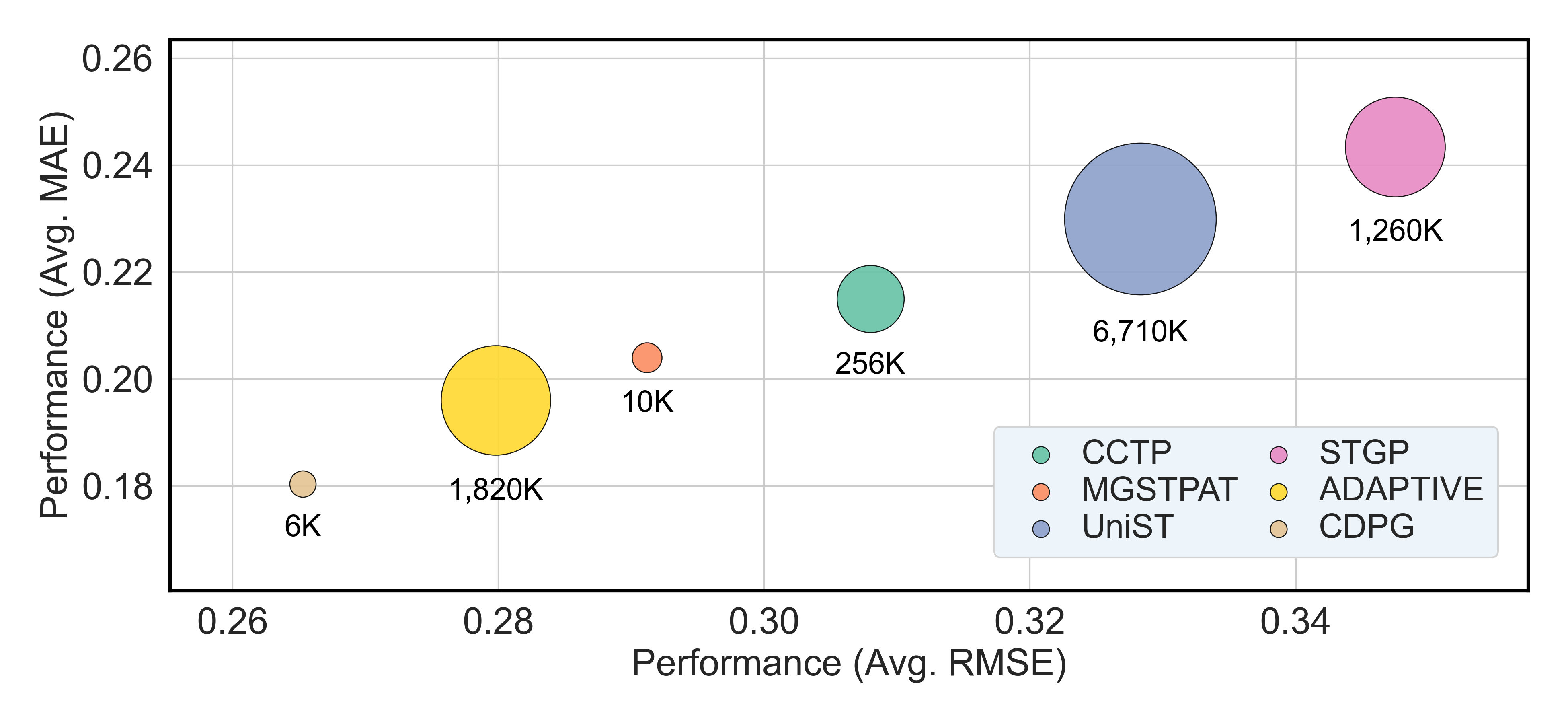}
\vspace{-2mm}
\Description{Comparison of prediction performance versus deployment model size for 1-step-ahead single-station traffic prediction across three datasets. The bubble size is proportional to the model size of each method.}
\caption{Comparison of prediction performance versus deployment model size for 1-step-ahead single-station traffic prediction across three datasets. The bubble size is proportional to the model size of each method.}
\label{fig:efficiency_bubble}
\vspace{-3mm}
\end{figure}

We analyze the computational costs to highlight the deployment advantages of CDPG, focusing on the resource requirements for per-station traffic prediction. Although our framework involves a generative process with a large parameter count during the parameter generation phase (10,390K parameters for the diffusion model), this design decouples parameter generation from real-time inference. Baseline methods like UniST (6,710K) and ADAPTIVE (1,820K) require deploying the entire model or generator to the target environment, even for predicting a single base station's value. 
\revise{ %R3-C4
In contrast, the CDPG diffusion model is used only in the offline parameter generation stage. 
}
For deployment, CDPG instantiates a lightweight predictor for each base station, with only 6K parameters. As a result, when evaluating the memory footprint for prediction services on edge devices (e.g., base station controllers), CDPG significantly reduces overhead compared to foundation models, while maintaining superior accuracy. To illustrate this, \figurename~\ref{fig:efficiency_bubble} shows the deployment model size for a single station versus prediction performance, demonstrating that CDPG strikes the best balance between deployment cost and accuracy.

\revise{ % R1-C4, R2-C4
We further report the practical cost of CDPG. Source expert training and diffusion training are both offline procedures. On a workstation with four NVIDIA GeForce RTX 4090 GPUs, training all source experts takes about two hours with parallel training, and training the diffusion generator takes about one hour. During target-city deployment, generating parameters for all target base stations takes about ten minutes, and fine-tuning the generated predictor for one target base station with three days of target data takes about 30 seconds. After fine-tuning, online prediction uses only the lightweight 6K-parameter predictor and does not require the diffusion generator or any LLM call.

Despite its lightweight online deployment, CDPG still incurs offline costs for source-expert and diffusion-model training, relies on the availability of satellite imagery and semantic descriptions, and requires a small amount of target-city data for fine-tuning.
}

\subsection{Discussion and Limitations}
\label{sec:discussion_limitations}

\minorrevise{
The experiments use a fixed three-day target-city adaptation budget,
providing a common basis for comparison. However, this setting does not
reveal how performance varies with target-data availability. Shorter
histories may make temporal-pattern matching and predictor fine-tuning
less reliable. Longer histories may benefit both CDPG and competing
methods, potentially changing their relative performance. The reported
gains therefore apply to the evaluated three-day setting and should not
be directly extrapolated to other adaptation budgets.
}

\minorrevise{
The geographic and temporal coverage is also limited. Evaluation
involves three Chinese cities---Shanghai, Nanjing, and Nanchang---with
records collected in 2014, 2021, and 2023, respectively. Since each city
is observed in a different period, the design does not isolate city
effects from collection-period effects. In particular, Nanchang's
seven-day record limits the assessment of week-to-week variation,
seasonal patterns, and long-term robustness. The results support
transfer among the evaluated city datasets, while generalization
across countries, operators, and substantially different urban
structures remains unverified. Future work will examine different
adaptation budgets and broader datasets covering more regions,
operators, and longer observation periods.
}

%% file: Related.tex
\section{Related Work}
\label{sec:related_work}

\subsection{Network Traffic Prediction}
Early approaches to network traffic prediction, relying on statistical models like SARIMA~\cite{shu2005wireless} and SVR~\cite{hong2012application}, had difficulty capturing non-linear relationships and scaling to large networks. With the rise of deep learning, methods such as recurrent neural networks, including LSTM and GRU~\cite{zhu2020lstm, yu2022network}, proved effective at capturing temporal dependencies but overlooked the inherent spatial correlations in mobile traffic patterns. As a result, spatio-temporal models integrating convolutional neural networks (CNNs) and graph neural networks (GNNs)~\cite{zhang2018long, yao2021mvstgn} emerged, modeling both spatial and temporal dependencies. In recent years, Transformers have also been adapted for mobile traffic prediction, leveraging their ability to capture long-range spatio-temporal dependencies, as seen in models such as STD-Net~\cite{hu2022citywide}, GLSTTN~\cite{gu2023spatial}, and TFT~\cite{kougioumtzidis2025mobile}. Despite their effectiveness, these methods rely heavily on abundant, high-quality historical data, which is essential for capturing complex patterns and temporal dependencies. In data-scarce scenarios, they struggle to generalize, often leading to overfitting and poor performance due to insufficient data diversity.

\revise{ % R3-C5

\subsection{Few-Shot Cross-City Knowledge Transfer}
Few-shot cross-city knowledge transfer aims to transfer knowledge from data-abundant source cities to target cities for specific tasks, effectively addressing data scarcity. Techniques such as domain adaptation, transfer learning, and meta-learning have been widely adopted. Traditional methods rely on region matching or clustering to bridge data gaps, using techniques like K-means~\cite{zeng2020traffic} for grouping similar regions and transferring knowledge between clusters~\cite{zhang2019deep} or within cells~\cite{dridi2021transfer}. Models like STCNet~\cite{zhang2019deep} and MGSTPAT~\cite{mo2025knowledge} improve flexibility by incorporating external dependencies, such as Point-of-Interest (POI) distributions and adaptive clustering, in data-scarce scenarios. Later approaches focus on probabilistic alignment of data distributions to capture subtle spatio-temporal relationships~\cite{xu2021spectragan, sun2022appshot, hui2023large}. Frameworks like CCTP~\cite{wu2021deep} and ADAPTIVE~\cite{zhang2023deep} use GANs for adversarial alignment to mitigate data distribution shifts between source and target cities. More recent methods, inspired by foundation models and LLMs, pre-train on large-scale data and adapt to new scenarios with minimal samples through learnable prompts~\cite{cao2023tempo, hu2024prompt, yuan2024unist}. 

Continual and cross-domain spatio-temporal learning methods are also related to cross-city transfer. CMuST~\cite{yi2024get} studies continuous multi-task spatio-temporal learning by modeling shared and task-specific knowledge across evolving tasks. SynEVO~\cite{liu2025synevo} focuses on cross-domain spatio-temporal adaptation by evolving shared knowledge across sample groups and domains. These methods are effective for exploiting task/domain commonality from observed spatio-temporal data streams. However, their adaptation signals are still mainly derived from available observations, task prompts, or domain sample groups, which can be limited when the target city provides only a few noisy traffic samples and exhibits strong base-station-level heterogeneity. CDPG focuses on this data-scarce cross-city setting by incorporating observation-independent spatial context, including satellite imagery and LLM-derived functional semantics, and by generating lightweight predictor parameters tailored to each target-city base station.
}

\revise{ %R1-C2 R2-C3 R3-C5
\subsection{Learning in Parameter Space}
\label{sec:related_parameter_generation}
The paradigm of learning in parameter space originates from HyperNetworks~\cite{ha2016hypernetworks}, which use a secondary network to generate parameters for a primary model. This idea has been adopted in spatio-temporal prediction. For instance, ST-MetaNet~\cite{pan2019urban} generates node-specific parameters based on static spatial properties and dynamic temporal patterns, while GPT-ST~\cite{li2023gpt} uses customized parameter learners to capture dynamic spatio-temporal dependencies. Diffusion models have recently shown strong potential for neural parameter generation. p-diff~\cite{wang2024neural} demonstrates that high-performing neural parameters can be generated from random noise, but it does not explicitly consider spatio-temporal contexts. GPD~\cite{yuan2024spatio} further introduces diffusive neural network generation for spatio-temporal few-shot learning. However, GPD focuses on general spatio-temporal few-shot learning and mainly relies on domain-level or knowledge-graph-based conditions to guide parameter generation. Such conditions may be insufficient for capturing fine-grained physical details, such as diverse land-use characteristics and latent functional semantics, and for modeling heterogeneous temporal dynamics in mobile traffic. In contrast, CDPG introduces a multimodal conditioning strategy that fuses satellite imagery with LLM-generated descriptions to extract spatial semantics, while integrating clustering-based temporal patterns to explicitly model diverse traffic dynamics.
}

%% file: Ref.bib
@article{su2025gennet,
  title={GenNet: A generative AI-driven mobile network simulator for multi-objective network optimization with RL},
  author={Su, Weikang and Li, Tong and Huang, Wenzhen and Huang, Zihan and Li, Yong},
  journal={IEEE Transactions on Mobile Computing},
  year={2026},
  volume={25},
  number={5},
  pages={6472-6483}
}

@article{xiao2025llm,
  title={An LLM-enhanced conditional diffusion model for mobile traffic prediction},
  author={Xiao, Zhu and Wang, Rui and Bai, Jing and Li, Tong and Zhang, Shiyuan and Li, Keqin and Han, Zhu},
  journal={IEEE Communications Magazine},
  volume={63},
  number={9},
  pages={60--67},
  year={2025},
  publisher={IEEE}
}

@article{zhang2026lsdm,
  title={LSDM: LLM-enhanced spatio-temporal diffusion model for service-level mobile traffic prediction},
  author={Zhang, Shiyuan and Li, Tong and Xiao, Zhu and Du, Hongyang and Huang, Kaibin},
  journal={IEEE Transactions on Networking},
  year={2026},
  publisher={IEEE}
}

@article{li2024mobile,
  title={Mobile user traffic generation via multi-scale hierarchical GAN},
  author={Li, Tong and Hui, Shuodi and Zhang, Shiyuan and Wang, Huandong and Zhang, Yuheng and Hui, Pan and Jin, Depeng and Li, Yong},
  journal={ACM Transactions on Knowledge Discovery from Data},
  volume={18},
  number={8},
  pages={1--19},
  year={2024},
  publisher={ACM, New York, NY}
}

@article{huang2025smaug,
  title={SMAUG: Semantic-Enhanced Mobile App Usage Data Generation With LLM},
  author={Huang, Zihan and Li, Tong and Li, Yong},
  journal={IEEE Transactions on Services Computing},
  year={2025},
  publisher={IEEE}
}

@article{yang2023transforms,
  title={{KGDA}: A knowledge graph driven decomposition approach for cellular traffic prediction},
  author={Gong, Jiahui and Li, Tong and Wang, Huandong and Liu, Yu and Wang, Xing and Wang, Zhendong and Deng, Chao and Feng, Junlan and Jin, Depeng and Li, Yong},
  journal={ACM Transactions on Intelligent Systems and Technology},
  volume={15},
  number={6},
  pages={1--22},
  year={2024},
  publisher={ACM New York, NY}
}

@article{zhang2021graph,
  title={Graph neural network-driven traffic forecasting for the connected internet of vehicles},
  author={Zhang, Qin and Yu, Keping and Guo, Zhiwei and Garg, Sahil and Rodrigues, Joel JPC and Hassan, Mohammad Mehedi and Guizani, Mohsen},
  journal={IEEE Transactions on Network Science and Engineering},
  volume={9},
  number={5},
  pages={3015--3027},
  year={2021},
  publisher={IEEE}
}

@article{xie2024m,
  title={M-STL: Multi-Range Multi-Level Spatial-Temporal Learning Model for Network Traffic Prediction},
  author={Xie, Ruotian and Wen, Jigang and Chen, Xiaodi and Xie, Kun and Xiong, Naixue and Zhang, Dafang and Xie, Gaogang and Li, Kenli and others},
  journal={IEEE Transactions on Network Science and Engineering},
  volume={11},
  number={5},
  pages={4315--4329},
  year={2024},
  publisher={IEEE}
}

@article{li2019privacy,
  title={Privacy-preserving traffic monitoring with false report filtering via fog-assisted vehicular crowdsensing},
  author={Li, Meng and Zhu, Liehuang and Lin, Xiaodong},
  journal={{IEEE Transactions on Services Computing}},
  volume={14},
  number={6},
  pages={1902--1913},
  year={2019},
  publisher={IEEE}
}

@article{gong2025sttf,
  title={STTF: A spatiotemporal transformer framework for multi-task mobile network prediction},
  author={Gong, Jiahui and Liu, Yu and Li, Tong and Ding, Jingtao and Wang, Zhaocheng and Jin, Depeng},
  journal={IEEE Transactions on Mobile Computing},
  volume={24},
  number={5},
  pages={4072--4085},
  year={2025},
  publisher={IEEE}
}

@inproceedings{gong2023empowering,
  title={Empowering spatial knowledge graph for mobile traffic prediction},
  author={Gong, Jiahui and Liu, Yu and Li, Tong and Chai, Haoye and Wang, Xing and Feng, Junlan and Deng, Chao and Jin, Depeng and Li, Yong},
  booktitle={Proceedings of the 31st ACM International Conference on Advances in Geographic Information Systems},
  pages={1--11},
  year={2023}
}

@inproceedings{chai2024knowledge,
  title={Knowledge guided conditional diffusion model for controllable mobile traffic generation},
  author={Chai, Haoye and Li, Tong and Jiang, Fenyu and Zhang, Shiyuan and Li, Yong},
  booktitle={Companion Proceedings of the ACM Web Conference 2024},
  pages={851--854},
  year={2024}
}

@article{bega2020deepcog,
  title={A comprehensive survey of the key technologies and challenges surrounding vehicular ad hoc networks},
  author={Xia, Zhenchang and Wu, Jia and Wu, Libing and Chen, Yanjiao and Yang, Jian and Yu, Philip S},
  journal={ACM Transactions on Intelligent Systems and Technology (TIST)},
  volume={12},
  number={4},
  pages={1--30},
  year={2021},
  publisher={ACM New York, NY}
}

@inproceedings{gong2023scalable,
  title={Scalable digital twin system for mobile networks with generative AI},
  author={Gong, Jiahui and Yu, Qiaohong and Li, Tong and Liu, Haoqiang and Zhang, Jun and Fan, Hangyu and Jin, Depeng and Li, Yong},
  booktitle={Proceedings of the 21st annual international conference on mobile systems, applications and services},
  pages={610--611},
  year={2023}
}

@article{li2025generative,
  title={Generative AI empowered network digital twins: Architecture, technologies, and applications},
  author={Li, Tong and Long, Qingyue and Chai, Haoye and Zhang, Shiyuan and Jiang, Fenyu and Liu, Haoqiang and Huang, Wenzhen and Jin, Depeng and Li, Yong},
  journal={ACM Computing Surveys},
  volume={57},
  number={6},
  pages={1--43},
  year={2025},
  publisher={ACM, New York, NY}
}

@article{zhang2022cityneuro,
  title={CityNeuro: Towards location and time prediction for urban abnormal events},
  author={Zhang, Mingyang and Li, Tong and Hui, Pan},
  journal={IEEE Transactions on Knowledge and Data Engineering},
  volume={35},
  number={12},
  pages={12125--12138},
  year={2022},
  publisher={IEEE}
}

@article{liu2026multi,
  title={A multi-granularity self-guiding graph diffusion model for predicting private car activity},
  author={Liu, Bo and Li, Tong and Xiao, Zhu and Chen, Wenjie and Hei, Yongqiang and Ding, Jingtao and Chen, Guo},
  journal={IEEE Transactions on Intelligent Transportation Systems},
  year={2026},
  publisher={IEEE}
}

@inproceedings{zhang2018long,
    title={Long-term mobile traffic forecasting using deep spatio-temporal neural networks},
    author={Zhang, Chaoyun and Patras, Paul},
    booktitle={{Proceedings of the 18th International Symposium on Mobile Ad Hoc Networking and
  Computing}},
    pages={231--240},
    year={2018}
}

@article{aouedi2025deep,
  title={Deep learning on network traffic prediction: Recent advances, analysis, and future directions},
  author={Aouedi, Ons and Le, Van An and Piamrat, Kandaraj and Ji, Yusheng},
  journal={{ACM Computing Surveys}},
  volume={57},
  number={6},
  pages={1--37},
  year={2025},
  publisher={ACM New York, NY}
}

@article{wu2021deep,
  title={Deep transfer learning across cities for mobile traffic prediction},
  author={Wu, Qiong and He, Kaiwen and Chen, Xu and Yu, Shuai and Zhang, Junshan},
  journal={{IEEE/ACM Transactions on Networking}},
  volume={30},
  number={3},
  pages={1255--1267},
  year={2021},
  publisher={IEEE}
}

@article{mo2025knowledge,
  title={Knowledge Transfer Service for Multi-Granular Traffic Prediction in Smart Cities},
  author={Mo, Jiqian and Chen, Junyang and Gong, Zhiguo},
  journal={{IEEE Transactions on Services Computing}},
  year={2025},
  volume={18},
  number={4},
  pages={2188-2200},
  publisher={IEEE}
}

@inproceedings{chai2025uomo,
  title={Uomo: A universal model of mobile traffic forecasting for wireless network optimization},
  author={Chai, Haoye and Zhang, Shiyuan and Qi, Xiaoqian and Qiu, Baohua and Li, Yong},
  booktitle={{Proceedings of the 31st ACM SIGKDD Conference on Knowledge Discovery and Data Mining}},
  pages={4308--4319},
  year={2025}
}

@inproceedings{yuan2024unist,
  title={Unist: A prompt-empowered universal model for urban spatio-temporal prediction},
  author={Yuan, Yuan and Ding, Jingtao and Feng, Jie and Jin, Depeng and Li, Yong},
  booktitle={Proceedings of the 30th ACM SIGKDD Conference on Knowledge Discovery and Data Mining},
  pages={4095--4106},
  year={2024}
}

@article{shu2005wireless,
  title={Wireless traffic modeling and prediction using seasonal ARIMA models},
  author={Shu, Yantai and Yu, Minfang and Yang, Oliver and Liu, Jiakun and Feng, Huifang},
  journal={IEICE Transactions on Communications},
  volume={88},
  number={10},
  pages={3992--3999},
  year={2005},
  publisher={The Institute of Electronics, Information and Communication Engineers}
}

@article{zeng2020traffic,
  title={Traffic prediction of wireless cellular networks based on deep transfer learning and cross-domain data},
  author={Zeng, Qingtian and Sun, Qiang and Chen, Geng and Duan, Hua and Li, Chao and Song, Ge},
  journal={IEEE Access},
  volume={8},
  pages={172387--172397},
  year={2020},
  publisher={IEEE}
}

@inproceedings{zhang2023deep,
  title={Deep transfer learning for city-scale cellular traffic generation through urban knowledge graph},
  author={Zhang, Shiyuan and Li, Tong and Hui, Shuodi and Li, Guangyu and Liang, Yanping and Yu, Li and Jin, Depeng and Li, Yong},
  booktitle={{Proceedings of the 29th ACM SIGKDD Conference on Knowledge Discovery and Data Mining}},
  pages={4842--4851},
  year={2023}
}

@inproceedings{hu2024prompt,
  title={Prompt-based spatio-temporal graph transfer learning},
  author={Hu, Junfeng and Liu, Xu and Fan, Zhencheng and Yin, Yifang and Xiang, Shili and Ramasamy, Savitha and Zimmermann, Roger},
  booktitle={{Proceedings of the 33rd ACM International Conference on Information and Knowledge Management}},
  pages={890--899},
  year={2024}
}

@article{li2022dmgan,
  title={DMGAN: Dynamic multi-hop graph attention network for traffic forecasting},
  author={Li, Rui and Zhang, Fan and Li, Tong and Zhang, Ning and Zhang, Tingting},
  journal={IEEE Transactions on Knowledge and Data Engineering},
  volume={35},
  number={9},
  pages={9088--9101},
  year={2022},
  publisher={IEEE}
}

@article{liu2024deep,
  title={Deep generative model and its applications in efficient wireless network management: A tutorial and case study},
  author={Liu, Yinqiu and Du, Hongyang and Niyato, Dusit and Kang, Jiawen and Xiong, Zehui and Kim, Dong In and Jamalipour, Abbas},
  journal={{IEEE Wireless Communications}},
  volume={31},
  number={4},
  pages={199--207},
  year={2024},
  publisher={IEEE}
}

@inproceedings{peebles2023scalable,
  title={Scalable diffusion models with transformers},
  author={Peebles, William and Xie, Saining},
  booktitle={{Proceedings of the IEEE/CVF International Conference on Computer Vision}},
  pages={4195--4205},
  year={2023}
}

@inproceedings{radford2021learning,
  title={Learning transferable visual models from natural language supervision},
  author={Radford, Alec and Kim, Jong Wook and Hallacy, Chris and Ramesh, Aditya and Goh, Gabriel and Agarwal, Sandhini and Sastry, Girish and Askell, Amanda and Mishkin, Pamela and Clark, Jack and others},
  booktitle={{International Conference on Machine Learning}},
  pages={8748--8763},
  year={2021}
}

@article{michel2019sixteen,
  title={Are sixteen heads really better than one?},
  author={Michel, Paul and Levy, Omer and Neubig, Graham},
  journal={{Advances in Neural Information Processing Systems}},
  volume={32},
  year={2019}
}

@inproceedings{yuan2024spatio,
  title={Spatio-Temporal Few-Shot Learning via Diffusive Neural Network Generation},
  author={Yuan, Yuan and Shao, Chenyang and Ding, Jingtao and Jin, Depeng and Li, Yong},
  booktitle={{The Twelfth International Conference on Learning Representations}},
year={2024}
}

@article{dosovitskiy2020image,
  title={An image is worth 16x16 words: Transformers for image recognition at scale},
  author={Dosovitskiy, Alexey},
  journal={{International Conference on Learning Representations}},
  year={2021}
}

@inproceedings{devlin2019bert,
  title={Bert: Pre-training of deep bidirectional transformers for language understanding},
  author={Devlin, Jacob and Chang, Ming-Wei and Lee, Kenton and Toutanova, Kristina},
  booktitle={{Proceedings of the 2019 conference of the North American chapter of the association for computational linguistics: human language technologies}},
  pages={4171--4186},
  year={2019}
}

@article{petitjean2011global,
  title={A global averaging method for dynamic time warping, with applications to clustering},
  author={Petitjean, Fran{\c{c}}ois and Ketterlin, Alain and Gan{\c{c}}arski, Pierre},
  journal={{Pattern recognition}},
  volume={44},
  number={3},
  pages={678--693},
  year={2011},
  publisher={Elsevier}
}

@inproceedings{dietterich2000ensemble,
  title={Ensemble methods in machine learning},
  author={Dietterich, Thomas G},
  booktitle={{International Workshop on Multiple Classifier Systems}},
  pages={1--15},
  year={2000},
  organization={Springer}
}

@article{hong2012application,
  title={Application of seasonal SVR with chaotic immune algorithm in traffic flow forecasting},
  author={Hong, Wei-Chiang},
  journal={{Neural Computing and Applications}},
  volume={21},
  number={3},
  pages={583--593},
  year={2012},
  publisher={Springer}
}

@article{gu2023spatial,
  title={A spatial-temporal transformer network for city-level cellular traffic analysis and prediction},
  author={Gu, Bo and Zhan, Junhui and Gong, Shimin and Liu, Wanquan and Su, Zhou and Guizani, Mohsen},
  journal={IEEE Transactions on Wireless Communications},
  volume={22},
  number={12},
  pages={9412--9423},
  year={2023},
  publisher={IEEE}
}

@article{kougioumtzidis2025mobile,
  title={Mobile Network Traffic Prediction Using Temporal Fusion Transformer},
  author={Kougioumtzidis, Georgios and Poulkov, Vladimir K and Lazaridis, Pavlos I and Zaharis, Zaharias D},
  journal={IEEE Transactions on Artificial Intelligence},
  year={2025},
  publisher={IEEE}
}

@article{zhang2019deep,
  title={Deep transfer learning for intelligent cellular traffic prediction based on cross-domain big data},
  author={Zhang, Chuanting and Zhang, Haixia and Qiao, Jingping and Yuan, Dongfeng and Zhang, Minggao},
  journal={IEEE Journal on Selected Areas in Communications},
  volume={37},
  number={6},
  pages={1389--1401},
  year={2019},
  publisher={IEEE}
}

@inproceedings{dridi2021transfer,
  title={Transfer learning for classification and prediction of time series for next generation networks},
  author={Dridi, Aicha and Afifi, Hossam and Moungla, Hassine and Boucetta, Ch{\'e}rifa},
  booktitle={ICC 2021 - IEEE International Conference on Communications}, 
  year={2021},
  volume={},
  number={},
  pages={1-6},
}

@article{wang2024neural,
  title={Neural network diffusion},
  author={Wang, Kai and Tang, Dongwen and Zeng, Boya and Yin, Yida and Xu, Zhaopan and Zhou, Yukun and Zang, Zelin and Darrell, Trevor and Liu, Zhuang and You, Yang},
  journal={arXiv preprint arXiv:2402.13144},
  year={2024}
}

@article{ha2016hypernetworks,
  title={Hypernetworks},
  author={Ha, David and Dai, Andrew and Le, Quoc V},
  journal={International Conference on Learning Representations},
  year={2017}
}

@inproceedings{alaluf2022hyperstyle,
  title={Hyperstyle: Stylegan inversion with hypernetworks for real image editing},
  author={Alaluf, Yuval and Tov, Omer and Mokady, Ron and Gal, Rinon and Bermano, Amit},
  booktitle={Proceedings of the IEEE/CVF Conference on Computer Vision and Pattern Recognition},
  pages={18511--18521},
  year={2022}
}

@article{li2023gpt,
  title={GPT-ST: generative pre-training of spatio-temporal graph neural networks},
  author={Li, Zhonghang and Xia, Lianghao and Xu, Yong and Huang, Chao},
  journal={Advances in Neural Information Processing Systems},
  volume={36},
  pages={70229--70246},
  year={2023}
}

@article{pan2020spatio,
  title={Spatio-temporal meta learning for urban traffic prediction},
  author={Pan, Zheyi and Zhang, Wentao and Liang, Yuxuan and Zhang, Weinan and Yu, Yong and Zhang, Junbo and Zheng, Yu},
  journal={IEEE Transactions on Knowledge and Data Engineering},
  volume={34},
  number={3},
  pages={1462--1476},
  year={2020},
  publisher={IEEE}
}

@inproceedings{pan2019urban,
  title={Urban traffic prediction from spatio-temporal data using deep meta learning},
  author={Pan, Zheyi and Liang, Yuxuan and Wang, Weifeng and Yu, Yong and Zheng, Yu and Zhang, Junbo},
  booktitle={Proceedings of the 25th ACM SIGKDD international conference on knowledge discovery \& data mining},
  pages={1720--1730},
  year={2019}
}

@inproceedings{zhu2020lstm,
  title={An LSTM-based traffic prediction algorithm with attention mechanism for satellite network},
  author={Zhu, Feiyue and Liu, Lixiang and Lin, Teng},
  booktitle={Proceedings of the 2020 3rd International Conference on Artificial Intelligence and Pattern Recognition},
  pages={205--209},
  year={2020}
}

@article{yao2021mvstgn,
  title={MVSTGN: A multi-view spatial-temporal graph network for cellular traffic prediction},
  author={Yao, Yang and Gu, Bo and Su, Zhou and Guizani, Mohsen},
  journal={IEEE Transactions on Mobile Computing},
  volume={22},
  number={5},
  pages={2837--2849},
  year={2021},
  publisher={IEEE}
}

@article{liu2026llmapp,
  title={LLMApp: Unleashing the Power of Large Language Model for Mobile App Usage Prediction},
  author={Liu, Bo and Li, Tong and Xiao, Miao and Xia, Beihao and Xiao, Zhu and Tang, Zhuo and Li, Kenli},
  journal={IEEE Transactions on Services Computing},
  year={2026},
  publisher={IEEE}
}

@inproceedings{yu2022network,
  title={Network traffic overload prediction with temporal graph attention convolutional networks},
  author={Yu, Qiaohong and Wang, Huandong and Li, Tong and Jin, Depeng and Wang, Xing and Zhu, Lin and Feng, Junlan and Deng, Chao},
  booktitle={2022 IEEE International Conference on Communications Workshops (ICC Workshops)},
  pages={885--890},
  year={2022},
  organization={IEEE}
}

@article{hu2022citywide,
  title={Citywide mobile traffic forecasting using spatial-temporal downsampling transformer neural networks},
  author={Hu, Yahui and Zhou, Yujiang and Song, Junping and Xu, Luyang and Zhou, Xu},
  journal={IEEE Transactions on Network and Service Management},
  volume={20},
  number={1},
  pages={152--165},
  year={2022},
  publisher={IEEE}
}

@article{cao2023tempo,
  title={TEMPO: Prompt-based Generative Pre-trained Transformer for Time Series Forecasting},
  author={Cao, Defu and Jia, Furong and Arik, Sercan O and Pfister, Tomas and Zheng, Yixiang and Ye, Wen and Liu, Yan},
  journal={The Twelfth International Conference on Learning Representations},
  year={2023}
}

@inproceedings{xu2021spectragan,
  title={SpectraGAN: Spectrum Based Generation of City Scale Spatiotemporal Mobile Network Traffic Data},
  author={Xu, Kai and Singh, Rajkarn and Fiore, Marco and Marina, Mahesh K and Bilen, Hakan and Usama, Muhammad and Benn, Howard and Ziemlicki, Cezary},
  booktitle={Proceedings of the 17th International Conference on Emerging Networking EXperiments and Technologies (CoNEXT '21)},
  pages={243--258},
  year={2021},
  publisher={Association for Computing Machinery},
  address={New York, NY, USA},
  doi={10.1145/3485983.3494838}
}

@article{sun2022appshot,
  title={Appshot: a conditional deep generative model for synthesizing service-level mobile traffic snapshots at city scale},
  author={Sun, Chuanhao and Xu, Kai and Fiore, Marco and Marina, Mahesh K and Wang, Yue and Ziemlicki, Cezary},
  journal={IEEE Transactions on Network and Service Management},
  volume={19},
  number={4},
  pages={4136--4150},
  year={2022},
  publisher={IEEE}
}

@inproceedings{hui2023large,
  title={Large-scale urban cellular traffic generation via knowledge-enhanced GANs with multi-periodic patterns},
  author={Hui, Shuodi and Wang, Huandong and Li, Tong and Yang, Xinghao and Wang, Xing and Feng, Junlan and Zhu, Lin and Deng, Chao and Hui, Pan and Jin, Depeng and others},
  booktitle={Proceedings of the 29th ACM SIGKDD Conference on Knowledge Discovery and Data Mining},
  pages={4195--4206},
  year={2023}
}

@inproceedings{huang2023safe,
  title={Safe-NORA: Safe reinforcement learning-based mobile network resource allocation for diverse user demands},
  author={Huang, Wenzhen and Li, Tong and Cao, Yuting and Lyu, Zhe and Liang, Yanping and Yu, Li and Jin, Depeng and Zhang, Junge and Li, Yong},
  booktitle={Proceedings of the 32nd ACM International Conference on Information and Knowledge Management},
  pages={885--894},
    publisher = {Association for Computing Machinery},
    address = {New York, NY, USA},
location = {Birmingham, United Kingdom},
  year={2023}
}

@article{sun2024dynamic,
  title={Deep reinforcement learning for vehicular edge computing: An intelligent offloading system},
  author={Ning, Zhaolong and Dong, Peiran and Wang, Xiaojie and Rodrigues, Joel JPC and Xia, Feng},
  journal={ACM Transactions on Intelligent Systems and Technology (TIST)},
  volume={10},
  number={6},
  pages={1--24},
  year={2019},
  publisher={ACM New York, NY, USA}
}

@article{li2023carbon,
  title={Carbon emissions of {5G} mobile networks in China},
  author={Li, Tong and Yu, Li and Ma, Yibo and Duan, Tong and Huang, Wenzhen and Zhou, Yan and Jin, Depeng and Li, Yong and Jiang, Tao},
  journal={Nature Sustainability},
  volume={6},
  number={12},
  pages={1620--1631},
  year={2023},
  publisher={Nature Publishing Group UK London}
}

@article{croitoru2023diffusion,
  title={Diffusion models in vision: A survey},
  author={Croitoru, Florinel-Alin and Hondru, Vlad and Ionescu, Radu Tudor and Shah, Mubarak},
  journal={IEEE Transactions on Pattern Analysis and Machine Intelligence},
  volume={45},
  number={9},
  pages={10850--10869},
  year={2023},
  publisher={Ieee}
}

@inproceedings{sohl2015deep,
  title={Deep unsupervised learning using nonequilibrium thermodynamics},
  author={Sohl-Dickstein, Jascha and Weiss, Eric and Maheswaranathan, Niru and Ganguli, Surya},
  booktitle={International Conference on Machine Learning},
  pages={2256-2265},
  year={2015}
}

@article{bai2025qwen3,
  title={Qwen3-vl technical report},
  author={Bai, Shuai and Cai, Yuxuan and Chen, Ruizhe and Chen, Keqin and Chen, Xionghui and Cheng, Zesen and Deng, Lianghao and Ding, Wei and Gao, Chang and Ge, Chunjiang and others},
  journal={arXiv preprint arXiv:2511.21631},
  year={2025}
}

@article{hurst2024gpt,
  title={Gpt-4o system card},
  author={Hurst, Aaron and Lerer, Adam and Goucher, Adam P and Perelman, Adam and Ramesh, Aditya and Clark, Aidan and Ostrow, AJ and Welihinda, Akila and Hayes, Alan and Radford, Alec and others},
  journal={arXiv preprint arXiv:2410.21276},
  year={2024}
}

@article{dubey2024llama,
  title={The llama 3 herd of models},
  author={Dubey, Abhimanyu and Jauhri, Abhinav and Pandey, Abhinav and Kadian, Abhishek and Al-Dahle, Ahmad and Letman, Aiesha and Mathur, Akhil and Schelten, Alan and Yang, Amy and Fan, Angela and others},
  journal={arXiv e-prints},
  pages={arXiv--2407},
  year={2024}
}

@article{wang2025capabilities,
  title={The fragile intelligence of GPT-5 in medicine},
  author={Handler, Rebecca and Sharma, Sonali and Hernandez-Boussard, Tina},
  journal={Nature Medicine},
  pages={1-3},
  year={2025},
  publisher={Nature Publishing Group US New York}
}

@inproceedings{jia2021scaling,
  title={Scaling up visual and vision-language representation learning with noisy text supervision},
  author={Jia, Chao and Yang, Yinfei and Xia, Ye and Chen, Yi-Ting and Parekh, Zarana and Pham, Hieu and Le, Quoc and Sung, Yun-Hsuan and Li, Zhen and Duerig, Tom},
  booktitle={International Conference on Machine Learning},
  pages={4904--4916},
  year={2021}
}

@article{li2021align,
  title={Align before fuse: Vision and language representation learning with momentum distillation},
  author={Li, Junnan and Selvaraju, Ramprasaath and Gotmare, Akhilesh and Joty, Shafiq and Xiong, Caiming and Hoi, Steven Chu Hong},
  journal={Advances in Neural Information Processing Systems},
  volume={34},
  pages={9694--9705},
  year={2021}
}

@inproceedings{yao2019learning,
  title={Learning from multiple cities: A meta-learning approach for spatial-temporal prediction},
  author={Yao, Huaxiu and Liu, Yiding and Wei, Ying and Tang, Xianfeng and Li, Zhenhui},
  booktitle={The world wide web conference},
  pages={2181--2191},
  year={2019}
}

@inproceedings{cirstea2022towards,
  title={Towards spatio-temporal aware traffic time series forecasting},
  author={Cirstea, Razvan-Gabriel and Yang, Bin and Guo, Chenjuan and Kieu, Tung and Pan, Shirui},
  booktitle={2022 IEEE 38th International Conference on Data Engineering (ICDE)},
  pages={2900--2913},
  year={2022},
  organization={IEEE}
}

@article{liu2025synevo,
  title={SynEVO: A neuro-inspired spatiotemporal evolutional framework for cross-domain adaptation},
  author={Liu, Jiayue and Yi, Zhongchao and Zhou, Zhengyang and Huang, Qihe and Yang, Kuo and Wang, Xu and Wang, Yang},
  journal={arXiv preprint arXiv:2505.16080},
  year={2025}
}

@article{yi2024get,
  title={Get rid of isolation: A continuous multi-task spatio-temporal learning framework},
  author={Yi, Zhongchao and Zhou, Zhengyang and Huang, Qihe and Chen, Yanjiang and Yu, Liheng and Wang, Xu and Wang, Yang},
  journal={Advances in Neural Information Processing Systems},
  volume={37},
  pages={136701--136726},
  year={2024}
}
